\documentclass[10pt,journal,compsoc]{IEEEtran}

\ifCLASSOPTIONcompsoc
  \usepackage[nocompress]{cite}
\else
  \usepackage{cite}
\fi

\ifCLASSOPTIONcompsoc
	\usepackage[caption=false,font=normalsize,labelfon t=sf,textfont=sf]{subfig}
\else
	\usepackage[caption=false,font=footnotesize]{subfig}
\fi
\usepackage[switch]{lineno}
\usepackage{xspace}
\usepackage{enumitem}
\usepackage{ragged2e}
\usepackage[hyphens]{url}
\usepackage[pagebackref,breaklinks,colorlinks]{hyperref}

\usepackage{amsmath,amsfonts}
\usepackage{amssymb}
\usepackage{bm}
\usepackage{bbm}
\usepackage{mathtools}
\usepackage{resizegather}

\usepackage{booktabs}
\usepackage[table]{xcolor}
\usepackage{multirow}
\usepackage{array}
\usepackage{longtable}
\usepackage{makecell}

\usepackage{graphicx}
\usepackage{stfloats}
\usepackage{subcaption}
\usepackage{caption}
\usepackage{rotating, float, caption}

\usepackage{algorithmic}
\usepackage{algorithm}
\usepackage{verbatim}
\usepackage{listings}

\begin{document}
\title{GeRe: Towards Efficient Anti-Forgetting in Continual Learning of LLM via General Samples Replay}

\author{
Yunan Zhang, Shuoran Jiang, Mengchen Zhao, Yuefeng Li, Yang Fan, Xiangping Wu, Qingcai Chen  

\IEEEcompsocitemizethanks{
\IEEEcompsocthanksitem Yunan Zhang, Shuoran Jiang, Yang Fan, Mengchen Zhao, Xiangping Wu, Qingcai Chen are with the Department of Computer Science and Technology, Harbin Institute of Technology, Shenzhen, China. (E-mail: zhangyunan@stu.hit.edu.cn, shuoran.chiang@gmail.com, yfan@stu.hit.edu.cn, zhaomengchen@stu.hit.edu.cn, wxpleduole@gmail.com,  qingcai.chen@hit.edu.cn)
\IEEEcompsocthanksitem Yuefeng Li is with Ysstech Info-Tech Co.,Ltd, Shenzhen.
\IEEEcompsocthanksitem Qingcai Chen and Xiangping Wu are the corresponding authors. (E-mail: wxpleduole@gmail.com, qingcai.chen@hit.edu.cn)
\IEEEcompsocthanksitem Code and Data Website: \url{https://github.com/Qznan/GeRe}
}

}

\markboth{Journal of \LaTeX\ Class Files,~Vol.~14, No.~8, August~2021}%
{Shell \MakeLowercase{\textit{et al.}}: A Sample Article Using IEEEtran.cls for IEEE Journals}

\maketitle

\begin{abstract}
The continual learning capability of large language models (LLMs) is crucial for advancing artificial general intelligence. However, continual fine-tuning LLMs across various domains often suffers from catastrophic forgetting, characterized by: 1) significant forgetting of their general capabilities, and 2) sharp performance declines in previously learned tasks.
To simultaneously address both issues in a simple yet stable manner, we propose General Sample Replay (GeRe), a framework that use usual pretraining texts for efficient anti-forgetting. Beyond revisiting the most prevalent replay-based practices under GeRe, we further leverage neural states to introduce a enhanced activation states constrained optimization method using threshold-based margin (TM) loss, which maintains activation state consistency during replay learning.
We are the first to validate that a small, fixed set of pre-collected general replay samples is sufficient to resolve both concerns---retaining general capabilities while promoting overall performance across sequential tasks.
Indeed, the former can inherently facilitate the latter.
Through controlled experiments, we systematically compare TM with different replay strategies under the GeRe framework, including vanilla label fitting, logit imitation via KL divergence and feature imitation via L1/L2 losses. Results demonstrate that TM consistently improves performance and exhibits better robustness. Our work paves the way for efficient replay of LLMs for the future.
Our code and data are available at \url{https://github.com/Qznan/GeRe}.

\end{abstract}


\IEEEdisplaynontitleabstractindextext
\IEEEpeerreviewmaketitle


\section{Introduction}

\IEEEPARstart{C}{ontinual} learning (CL) of large language models (LLMs) remains challenging for real-world applications. For instance, continual finetuning often degrades general capabilities, particularly over long task sequences. The finetuned model forgets its original world knowledge or basic instruction-following skills~\cite{luo2023empirical,zheng2024concept}. Additionally, the overall performance on sequential downstream tasks often deteriorates due to forgetting of previously learned tasks, caused by inter-task conflicts.
This phenomenon, also known as catastrophic forgetting, often compels practitioners to seek complex CL solutions. However, the contemporary LLM system, marked by architectural bulkiness and computational heaviness, is imperative to call for a simple yet stable approach to effectively mitigate forgetting.
In this context, our research aims to review and develop an efficient and general anti-forgetting method of CL adapted to the LLM era.

\begin{figure}[tbp]
    \centering
    \includegraphics[width=0.5\textwidth]{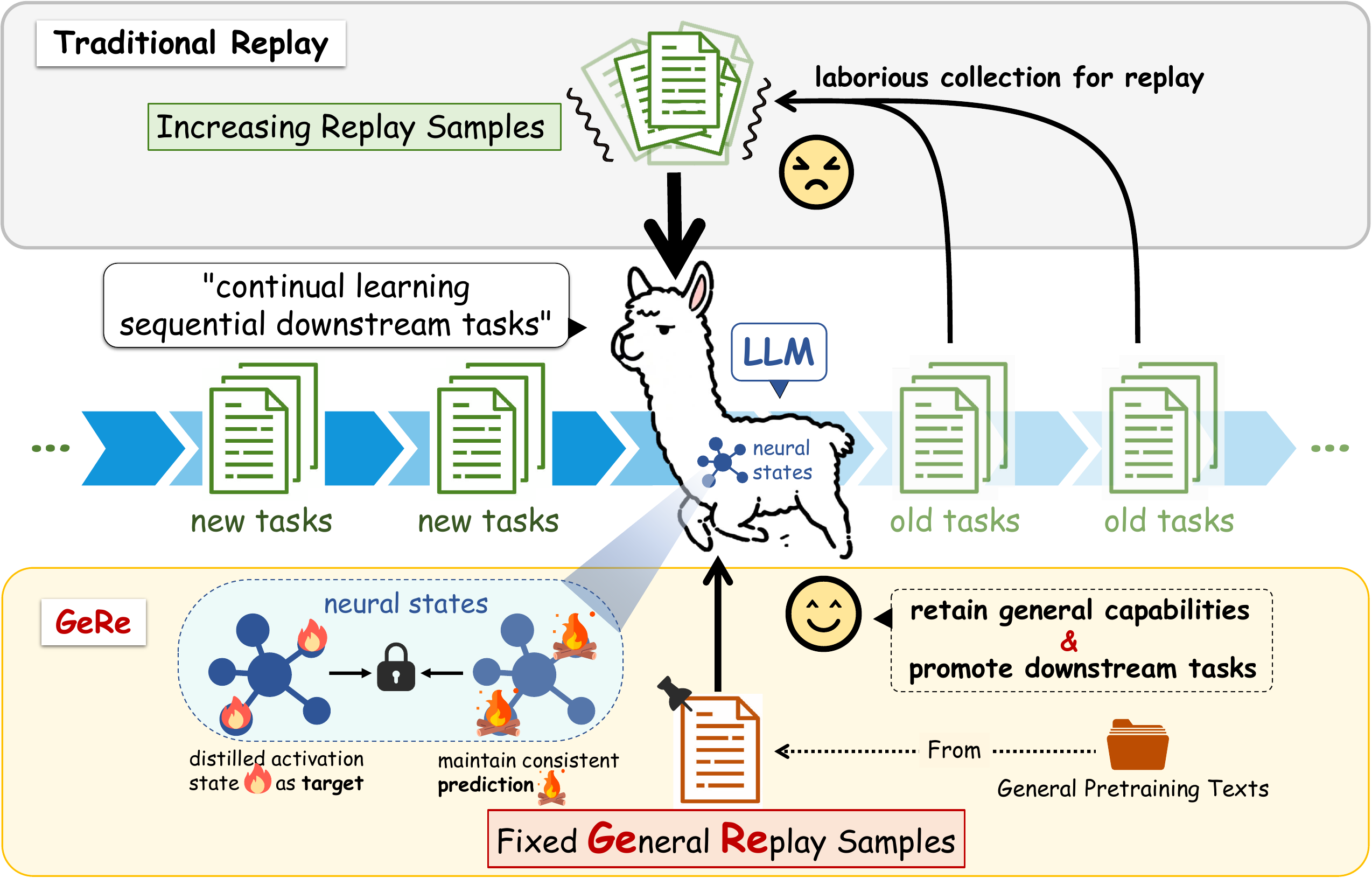}
    \caption{
    Traditional replay vs. GeRe: unlike traditional replay requiring laborious collection of an increasing set of downstream replay samples, GeRe simply employs a fixed set of general replay samples to not only retain general capabilities in continual learning, but also enhance the overall performance of learned downstream tasks. The blue oval is the threshold-based margin loss that imposes consistency constraint on neural activation state under GeRe frameworks.
    }
    \label{fig:intro}
\end{figure}

Historically, solutions for CL are primarily categorized into three traditional branches: replay-based, regularization-based, and architecture-based methods~\cite{shi2024continual}.
Considering the massive number of parameters in LLMs and their widely accepted fixed structures, it appears prohibitive and impractical to regularize all parameters or frequently expand the architecture for every new task. 
Thus current practice in LLMs continual learning regularly prioritizes replay-based methods due to its simplicity.
For instance, practitioners commonly mix a certain proportion of general task samples during finetuning for downstream tasks \cite{touvron2023llama,yang2024qwen2}. However, the underlying mechanisms and optimal strategies of these replay-based methods tailored for LLMs remain insufficiently explored and analyzed.

In this work, we
(1) systematically revisit the replay mechanisms targeting LLMs under a newly introduced general sample replay (GeRe) framework and,
(2) present a threshold-based margin (TM) loss for activation state constrained optimization.
Specifically, we prepares a fixed, permanently reusable set of general replay samples (e.g., the commonly used pretraining texts) and leverages the TM loss to maintain consistent neuron activation states, ultimately resisting various forms of forgetting.

The approach is motivated by two ideas:
(1) From a cognitive perspective, a learner obtaining superior general capabilities is more likely to achieve better generalization and robustness in downstream tasks. Leveraging its comprehensive knowledge, such a learner can reduce conflicts arising from overfitting to specific tasks, thereby mitigating task forgetting. Consequently, it is worth exploring how to utilize general replay samples to retain general capabilities.
(2) In the human brain, critical information is sparsely distributed across a few activated neurons \cite{olshausen1996emergence,wolfe2010sparse}.
Therefore, in replay-based continual learning (replay learning), the activation states of neurons evoked by replay samples may require deeper attention. By designing an activation state constrained optimization, we seek a less rigid but more informative target that enables the replay learning to be more robust and generalizable.

Through the paper we have explored and answered 2 pressing questions in real-world LLMs continual learning scenario:

\noindent
\textbf{Q1: Can we simply select a fixed set of replay samples once and for all?}
To retain general capabilities, contemporary strategies for mixing replay samples in LLM training may be as laborious as feature engineering, requiring careful selection of both the proper size and specific replay samples tailored to the particular downstream task.
For instance, even with a fixed mixing ratio, we still need to frequently resize the replay samples set and select an appropriate subset or superset to adapt to the varying data scale of incoming tasks.
For this question, we have empirically validated that constructing a fixed set of randomly selected general replay samples (e.g., 1k texts from the widely available general pretraining corpus) can be durably applied to fulfill all replay needs in subsequent tasks, while successfully preserving general capabilities.
This becomes more pronounced when integrating replay with feature-based distillation, as it fully exploits information from these limited replay samples rather than merely fitting their explicit labels.
To our knowledge, we are the first to propose that a fixed set of general replay samples can efficiently adapt to real-world continual learning scenario involving long sequences of tasks under full or LoRA settings, which holds significant practical implications.

\noindent
\textbf{Q2: Can general replay samples alone facilitate continual learning in sequential downstream tasks, typically without any of task replay samples?}
Normally, collecting task replay samples from each old task in subsequent learning is necessary to maintain their long-term performance. However, we believe that the learning  efficacy of any downstream task fundamentally depends on the LLM's general knowledge. For this question, we have encouragingly validated that the aforementioned fixed set of general replay samples, under our optimization approach, can effectively promote the persistent retention of previously learned task knowledge, mitigating the forgetting induces by inter-task conflicts.
The results demonstrate the feasibility of conveniently utilizing only predetermined general replay samples to resist task-specific forgetting in future applications.

These answers highlight the advantages of the proposed GeRe framework. Furthermore, we enhance feature-based replay learning under GeRe by introducing activation state constrained optimization, which statistically determines activation states and optimizes using a threshold-based margin loss. This relatively lightweight constraint on feature values empirically exhibits better robustness and generalizability compared to the conventional yet rigid L1/L2 fitting manner.

Our contributions are as follows:
\begin{itemize}[leftmargin=*, align=left]

\item \textbf{1. \textit{GeRe}}: We first demonstrate that a fixed set of predefined general replay samples can be reused throughout the entire continual finetuning process, effectively preserving LLM's original general capabilities. Crucially, replaying any downstream task sample proves unnecessary, as maintaining general capabilities alone enhances overall downstream tasks performance.

\item \textbf{2. \textit{TM loss}}: We pioneer a comprehensive comparison of commonly used replay-based practices for continual learning in LLMs, exploring their integration with various knowledge distillation strategies. Among these, our proposed threshold-based margin loss, motivated by the previously overlooked activation state constraint, achieves SoTA performance.

\item \textbf{3.}
Our method shows robustness not only to learning rate---a critical hyperparameter seriously impacting both knowledge updating and retention---but also to intrinsic optimization dynamics, as evidenced by optimization landscape visualization, highlighting its practical utility.

\end{itemize}

\section{Related Works}\label{sec:Related Works}

\subsection{Continual Learning}
Continual learning, also known as lifelong or incremental learning, refers to the ability of a machine learning model to learn from a stream of data over time, while retaining knowledge from previous tasks and adapting to new ones without forgetting~\cite{parisi2019continual}. Unlike traditional learning paradigms, where models are trained on static datasets, continual learning should addresses the dynamic nature of real-world applications, where data distributions and tasks evolve over time.

A central challenge in continual learning is the catastrophic forgetting problem, where a model tends to forget previously learned knowledge when trained on new tasks \cite{french1999catastrophic}. To mitigate this, various strategies have been proposed, including replay-based, regularization-based, and architecture-based methods. For instance, replay-based methods like Experience Replay \cite{rolnick2019experience} store and replay samples from past tasks to maintain performance, while regularization-based methods like Elastic Weight Consolidation (EWC) \cite{kirkpatrick2017overcoming} introduce a regularization term to preserve important parameters for previous tasks. Similarly, Learning without Forgetting (LwF) \cite{li2017learning} uses knowledge distillation to regularize the model by minimizing the divergence between its current and previous outputs. Architecture-based methods allocate distinct subsets of model parameters to different tasks to prevent interference. For example, Progressive Neural Networks (PNNs) \cite{rusu2016progressive} expand the network architecture by adding new layer of parameters for each task while freezing existing ones. In addition Mask-Based Methods \cite{mallya2018piggyback} learns task-specific masks to trigger or suppress parameters dynamically.

In the LLM era, continual pretraining or finetuning has become essential for model iteration and advancement. We focus on continual finetuning, where the most common practice involves replay-based methods, which typically incorporate general corpus to preserve the model's general capabilities for downstream tasks.
In this work, we explore a further integration with regularization-based techniques, i.e., the distillation strategy used in LwF. Specifically, we pre-generate and store the feature representations of the replay samples, which are then used as targets within a distillation framework to effectively leverage this information.
Moreover, we include the LoRA~\cite{hu2022lora} setup since it is widely adopted in finetuning due to its strong generalization and resistance to forgetting. It can be considered as another type of architecture-based method, as it typically trains only a small fraction of parameters.

\begin{figure*}[htbp]
    \centering
    \includegraphics[width=0.9\textwidth]{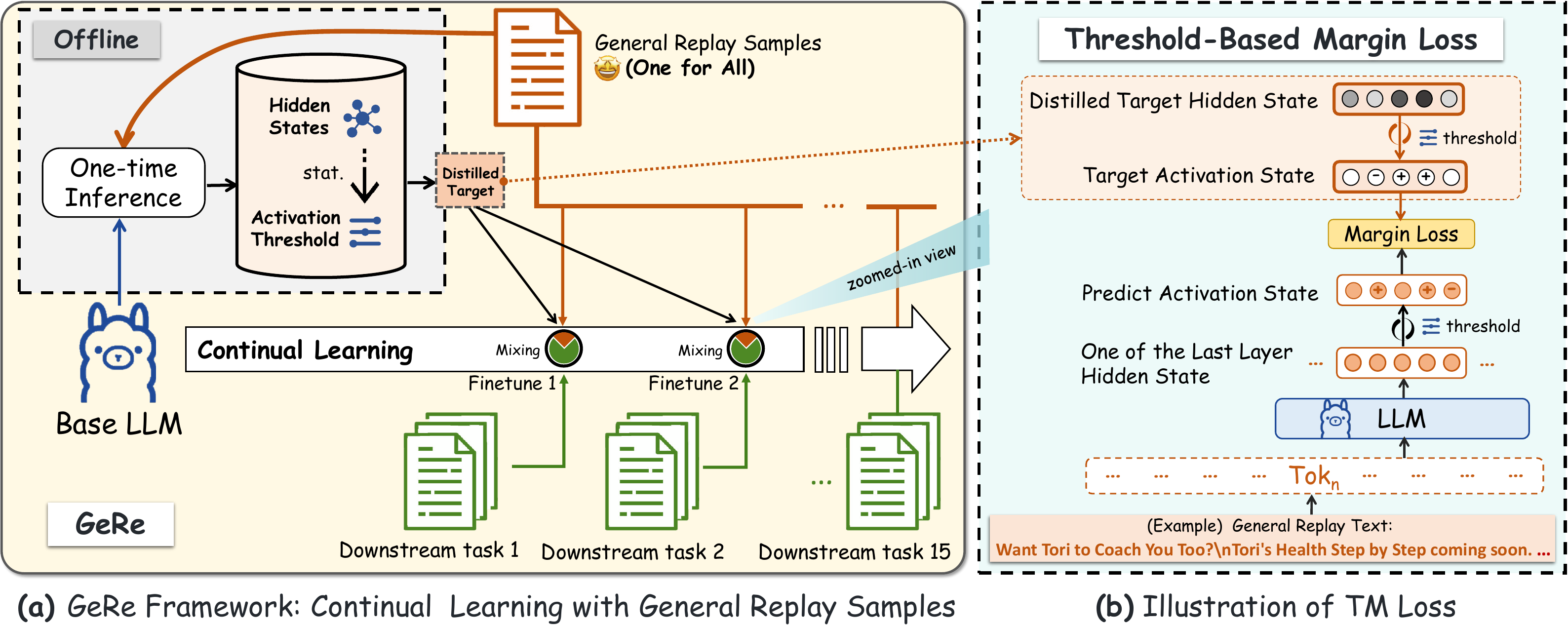}
    \caption{(a) Flowchart of the GeRe framework using general replay samples, including distillation of hidden states and the derived activation state in offline mode, and continual learning across sequential tasks with mixing general samples for replay.
    (b) Illustration of threshold-based margin loss, which transforms the hidden values into discrete activation states on both the target and prediction followed by margin loss calculation.
    }
    \label{fig:main_pic}
\end{figure*}

\subsection{Knowledge Distillation}
Knowledge distillation (KD)~\cite{hinton2015distilling} aim to compress large models into smaller, efficient ones by transferring knowledge from a teacher model to a student model. This process is achieved by minimizing the difference between their output distributions, where the student learns from the teacher's soft labels rather than the original dataset's hard labels~\cite{sanh2019distilbert,jiao2020tinybert,zhou2022distilling}.
KD can be implemented generally with two types: logit-based imitation and feature-based imitation~\cite{zhengetal2023ld}. The former involves matching the predictions and target distributions using the Kullback-Leibler divergence (KL) loss with a temperature-based softmax normalization, while the latter focuses on aligning the intermediate representations in the feature space through similarity-based functions. 

KD has already been applied in continual learning, with the key distinction lying in how the target and prediction are defined. Taking LwF for example, when the new task's samples arrived, the model preliminarily computed the logits of these samples at the output heads regarding old tasks, served as the distilled pseudo-targets.
During subsequent training, the real-time predicted logits at these old tasks output heads are constrained to match the precomputed pseudo-targets.
In this case, the teacher and student models are essentially the same model. This self-distillation mechanism~\cite{wang2023distill} enables the model to retain performance on previously learned tasks while adapting to new ones.

However, few research emphasizes similarity of feature in KD typically applied for continual learning of LLMs~\cite{xu2024survey}.
Our work thereby delves deeper into studying efficient mechanism combining replay and distillation methods, where labels or features of replay samples are pre-distilled to serve as pseudo-targets, enabling persistent fitting of replay samples during continual learning.
We empirically compare the effect of using replay samples simply versus leveraging replay samples under both logit-based imitation via KL divergence and feature-based imitation via L1 or L2 function. The study offers a comprehensive comparison of diverse replay strategies.
\section{Proposed Method}\label{sec:Proposed Method}
This section provides a detailed introduction to the overall process of GeRe framework and the proposed replay-based activation state constrained optimization.
As outlined in Fig.\ref{fig:main_pic}(a),
we first collect a small-scale set of general samples for permanently available replay.
Then these data are proactively distilled using the untuned base LLM to derive the activation threshold, which determines the activation state. Subsequently, continual finetuning is performed on a mixed data containing downstream task and general replay samples, jointly optimizing a specialized replay-based objective alongside the standard cross-entropy loss.
Fig.\ref{fig:main_pic}(b) illustrates our proposed threshold-based margin (TM) loss, which transforms the given optimization target into activate states through thresholds on both sides and employs a margin loss for constraint.
(The different optimization targets used by other competitors are shown in Fig. \ref{fig:compare}.)

\subsection{Distilled Activation States}\label{sec:Neural Activation Distribution}
In deep neural networks, neuron activation values refer to the outputs of each layer's sub-network.
Taking the Transformer-based LLMs as an example, the activation values refer to the output of the feed-forward network within each layer.
These activations are progressively passed through residual connections, evolving started from the input and ultimately forming the network's final output.

Analogous to the activation states of neurons in the human brain, we propose to categorize the neural network activations into three distinct states: positive activation, negative activation, and non-activation. These states exhibit discrete sparsity patterns while encoding specific semantic information.
Building on this, we hypothesize that during continual finetuning, the original activation states represented with general replay samples can effectively reflect the model's general capabilities. Therefore, in our replay learning, we employ feature-based imitation to utilize these activations as targets, thereby preserving the essential characteristics of the model's learned representations.

\subsubsection{Feature-Based Distillation}
Given a general replay sample set $\mathcal{D}^{\text{(g)}} = \{s_1, s_2, \dots, s_N\}$ comprising N natural sentences $s$, we feed these sentence samples into the base LLM, performing forward propagation to obtain the activation values (i.e., the hidden states output of each layer) as follows:
\begin{equation}
    \bar{\mathbf{h}} = \text{LLM}(s)
\end{equation}
where $\bar{\mathbf{h}}\in\mathbb{R}^{n^{t} \times n^{d} \times L}$ is activation value tensor,
$n^{\text{t}}$ is the length (number of tokens) of the input,
$n^{\text{d}}$ is the dimension of the hidden states,
$L$ is the number of layers in LLM.
We distill these feature-base activation values of all samples in $\mathcal{D}^{\text{(g)}}$ to form $\mathcal{H}^{\text{(g)}}=\{\bar{\mathbf{h}}^1,\bar{\mathbf{h}}^2,\dots,\bar{\mathbf{h}}^N\}$.

\subsubsection{Activation Threshold}
After distilling all the activation values of samples in $\mathcal{D}^{\text{(g)}}$,
we statistically determine the activation threshold and accordingly infer the activation state.
Specifically, for each activation value $\bar{h}_{j,k,l}$ corresponding to the $k$-th dimension of hidden state at the $l$-th layer, we compute its mean and variance across the entire $\mathcal{H}^{\text{(g)}}$ over the size $N$ and length $n^{\text{t}}$,
yielding $mean_l = (m_1,m_2,m_k,\dots,m_{n^{\text{d}}})_l$ and $std_l = (\sigma_1,\sigma_2,\sigma_k,\dots,\sigma_{n^{\text{d}}})_l$ relative to the $l$-th layer.
In practice, since the hidden state of the last layer encodes the majority of the semantic information for the model's final predictions, we choose to utilize only the last layer for subsequent computations, which serve as the constraint optimization objective.
Therefore, we assume $l\!=\!L$ (the last layer) and omit the subscript $l$ in all the following formulas, (e.g., $\bar{h}_{j,k}\!:=\!\bar{h}_{j,k,l=L}$). Each component $m_k$ and $\sigma_k$ is computed as follows:

\begin{gather}
    m_k =\frac{1}{N \times n^{\text{t}}} \sum_{i=1}^{N}\sum_{j=1}^{n^{\text{t}}}\bar{h}_{j,k}^{i}
    \\
    \sigma_k = \sqrt{ \frac{1}{N \times n^{\text{t}}} \sum_{i=1}^{N}\sum_{j=1}^{n^{\text{t}}} (\bar{h}_{j,k}^{i}-m_k)^2 }
\end{gather}
where $i$ ranges over the $\mathcal{D}^{\text{(g)}}$ size $N$ and $j$ ranges over the number of tokens within the current sample, $k$ denote the $k$-th dimension. We further utilize the characteristics of Gaussian distribution to define the activation thresholds, considering one standard deviation above the mean as the positive activation threshold: $\tau^+\!=\!m+1\sigma$, and one standard deviation below the mean as the negative activation threshold: $\tau^-\!=\!m-1\sigma$. Each $\bar{h}_k$ relative to the $k$-th dimension possesses two thresholds as follows:
\begin{gather}
    \tau_k = (\tau^-_k,\quad \tau^+_k) = (m_k-1\cdot\sigma_k,\quad m_k+1\cdot\sigma_k)
\end{gather}
and we define three types of activation state: 1) values greater than $\tau^+$ are considered positively activated, 2) values less than $\tau^-$ are considered negatively activated, 3) values between $\tau^-$ and $\tau^+$ are considered non-activated:
\begin{equation}
\text{state}_{k} =  
\begin{cases}  
\text{positively activated} & \text{if } \text{value} < \tau_k^- \\
\text{non-activated} & \text{if }  \tau_k^- \le \text{value} \le \tau_k^+\\
\text{negatively activated} & \text{if } \text{value} > \tau_k^+ \\
\end{cases}
\label{eq:state}
\end{equation}

According to Gaussian distribution, about 68.27\% of the activation values are considered non-activated, which aligns with the assumption that only a subset of neurons plays a critical role during forward propagation. 

Once the thresholds for the $\mathcal{D}^{\text{(g)}}$ are determined, they can be permanently applied to subsequent downstream task finetuning conveniently.
Notably, we can also preemptively transform the float-type activation values into binary-type activation states to reduce the storage overhead.

\subsection{Threshold-Based Margin Optimization}\label{sec:tm}
This section describes the computation process of the proposed TM loss.
Optionally we can randomly select a subset of samples from $\mathcal{D}^{\text{(g)}}$ for actually replay. However, if the original size is small, using the complete set is recommended.
Specifically, given the previously determined positive and negative activation thresholds, these samples are jointly optimized with the downstream task samples during continual finetuning. Detailed steps are as follows.

\subsubsection{Batch Insertion} \label{sec:BI}
Traditional replay-based methods simply mix replay samples with downstream task training samples randomly. However, it requires considering the scale of both samples set and thereby adjusting the mixing ratio. Additionally, during optimization, it is possible that a given batch may contain no replay samples, resulting in gradients that are exclusively influenced by the downstream task samples. To address this issue, we propose the Batch Insertion (BI) strategy, which ensures that a specific proportion of replay samples is included in each training batch. This strategy encourages the influence of the gradient update direction by the replay samples in every iteration, helping to retain the general capabilities of the LLM, meanwhile avoids the cumbersome adjustment of mixing ratios for datasets of varying scales.

Given the finetuning batch size $n^{\text{batch}}$, we define the Batch Insertion ratio as $\rho^{\text{BI}}$, indicating that $\rho^{\text{BI}} \times n^{\text{batch}}$ samples in each batch are replay samples. This can be easily implemented by modifying the \textit{Sampler} Class within \textit{Torch DataLoader}.

\subsubsection{Loss Calculation} 
During the joint training process, the proposed TM loss for the replay samples within each batch is computed as follows:  

\begin{equation}
\mathcal{L}^{\text{TM}}_{j,k} =  
\begin{cases}  
\max(\hat{h}^{j,k} - \tau_k^-, 0) & \text{if } \bar{h}^{j,k} < \tau_k^- \\  
\begin{aligned}  
&\max(\hat{h}^{j,k} - \tau_k^+, 0) \\  
&+ \max(\tau_k^- - \hat{h}^{j,k}, 0)  
\end{aligned} & \text{if } \tau_k^- \le \bar{h}^{j,k} \le \tau_k^+ \\  
\max(\tau_k^+ - \hat{h}^{j,k}, 0) & \text{if } \bar{h}^{j,k} > \tau_k^+ 
\end{cases}
\label{eq:tml}
\end{equation}
where $\mathcal{L}^{\text{TM}}_{j,k}$ denotes the TM loss for the $k$-th dimension of the hidden state on the $j$-th token (at the last layer), $\bar{h}$ is the precomputed target activation values while $\hat{h}$ is the currently predicted activation values.
This piecewise loss function guides the optimization direction of the predicted value when the target value resides in the negatively activated, non-activated, and positively activated states, respectively.
The overall TM loss for each replay sample sentence is computed as follows:
\begin{gather}
\mathcal{L}^{TM} = \frac{1}{n^{\text{t}} \times n^{\text{d}}} \sum_{j=1}^{n^{\text{t}}} \sum_{k=1}^{n^{\text{d}}} \mathcal{L}^{TM}_{j,k}
\end{gather}

\subsubsection{Dynamic Weight Balancing} \label{sec:DW}
During training, we jointly optimize the TM loss $\mathcal{L}^{\text{TM}}$ regarding general replay samples and the standard Cross-Entropy (CE) loss $\mathcal{L}^{\text{CE}}$ regarding downstream task samples.
To prevent the model from being overly biased toward optimizing either loss, we adopt a dynamic loss weighting strategy to balance their magnitudes as follows:
\begin{gather}
\omega^{\text{TM}} = {\text{detach}}(\mathcal{L}^{\text{CE}} / \mathcal{L}^{\text{TM}})
\label{eq:tml_weight}
\\
\mathcal{L} = \mathcal{L}^{\text{CE}} + \omega^{\text{TM}} \cdot \mathcal{L}^{\text{TM}}
\label{eq:loss_final}
\end{gather}
where $\mathcal{L}$ denotes the final total loss for continual finetuning. $\omega^{\text{TM}}$ is the dynamic weight to dynamically scale the magnitude of the TM loss to match that of the CE loss during joint optimization.
The ${\text{detach}}()$ function indicates that the weight value is detached from gradient backpropagation, preventing it from being optimized.
Notably, this approach is experimental, employing fixed or dynamic weights depending on practice.

\section{Experiments}\label{sec:Experiments}
In this section, we evaluate the performance of our proposed method using a representative base-LLM Llama-3.1-8B~\cite{grattafiori2024llama} along with 15 downstream tasks under continue learning regime. We first introduce the datasets, metrics and experimental settings, followed by detailed analyses of the experimental results and landscape visualization~\cite{li2018visualizing,wang2024comprehensive} exploring robustness.

\begin{table}[b]
\footnotesize
\centering
\caption{The statistic of the 15 downstream tasks.}
\label{tab:stats_15tasks}
\begin{tabular}{c|c|c|c}
\toprule
Task Type& Datasets & \# of Train & \# of Test\\
\midrule
SC & yelp & 5000 & 7600 \\
SC & amazon & 5000 & 7600 \\
NLI & MNLI & 3000 & 7600 \\
NLI & CB & 250 & 56 \\
COPA & COPA & 400 & 100 \\
QQP & QQP & 2000 & 7600 \\
NLI & RTE & 2000 & 277 \\
SC & IMDB & 2000 & 7600 \\
SC & SST-2 & 2000 & 872 \\
TC & dbpedia & 14000 & 7600 \\
TC & agnews & 4000 & 7600 \\
SC & yahoo & 10000 & 7600 \\
MultiRC & MultiRC & 2000 & 4848 \\
BoolQA & BoolQA & 2000 & 3270 \\
WiC & WiC & 2000 & 638 \\

\bottomrule
\end{tabular}

\end{table}

\subsection{Datasets}
For the general replay sample set $\mathcal{D}^{\text{(g)}}$, we randomly select 1K samples from the open-source SlimPajama-627B corpus~\cite{cerebras2023slimpajama}, which is a cleaned and deduplicated version of RedPajama~\cite{weber2024redpajama} that reproduces the collection of LLaMA training data.
We release the complete selected samples used throughout this paper to ensure reproducibility. Notably, the selection process is arbitrary rather than deliberately curated (see Appendix for details), which further substantiating the robustness and universality of our method with respect to the replay data.
This replay data potentially reflect the general ability of the base LLM model, which is used to calculate the activation threshold and to compute the threshold-based margin loss.

For the downstream finetuning tasks, we adopt a long-sequence continual learning benchmark comprising as many as 15 diverse datasets~\cite{razdaibiedina2023progressive}, which enables a comprehensive evaluation of model performance in practical scenarios under more demanding and challenging conditions.
The benchmark integrates 5 datasets (yelp, amazon, dbpedia, agnews, yahoo) from the standard CL benchmark~\cite{zhang2015character,qin2021lfpt5}, 4 datasets (MNLI, QQP, RTE, SST-2) from the GLUE benchmark~\cite{wang2018glue}, 5 datasets (CB, COPA, MultiRC, BoolQA, WiC) from the SuperGLUE benchmark~\cite{wang2019superglue}, and the IMDB movie reviews dataset~\cite{maas2011learning}.
In alignment with~\cite{razdaibiedina2023progressive}, we utilize the available validation set for each dataset as the test set since test data is not available.
However, unlike their setting, which randomly selects fixed number of training samples per dataset (i.e., potentially up-sampling or down-sampling), we employ the original full training set for each dataset to better align with real-world scenarios where the data quantity distribution across tasks is inherently imbalanced.
In continual finetuning, we train each task until the training loss converges without validation set.
We proceed to train the next task after the previous one is finish, and the training data for each task was no longer available once used.

Table~\ref{tab:stats_15tasks} presents the dataset statistics for the 15 tasks. Examples mainly including instructions, inputs, and golden answers for each dataset are provided in the Appendix.

\subsection{Metrics}
We evaluate the performance of the final model (i.e., after continual finetuning on 15 tasks) from two dimensions comprising general capabilities and the average accuracy over all downstream tasks.

For general capabilities, We employ MMLU~\cite{hendrycks2020measuring} benchmark, which spans 57 diverse disciplines ranging from STEM, humanities and social sciences, etc., to rigorously measure both factual knowledge and analytical skills across multiple levels of complexity. We use five-shot setting and discriminative evaluation.

For ability to effectively learn the sequential downstream tasks,
we assess the Average Performance (AP)~\cite{chaudhry2018riemannian} of the final model via obtaining task-wise accuracies and then computing their mean. AP reflects the model's overall performance across multiple tasks and its ability to retain knowledge from previously learned.
Notably, we also evaluated the multi-task learning (MTL) regime which finetunes on the combined dataset of all 15 tasks, serving as the theoretical upper bound performance for continual learning.

Finally, we compute the F1 average of the MMLU and AP to reflect the holistic performance of model in maintaining its original general capabilities while effectively learning downstream tasks.
All experimental results are reported as the average of 3 runs.

\begin{figure*}[htbp]
    \centering
    \includegraphics[width=0.8\textwidth]{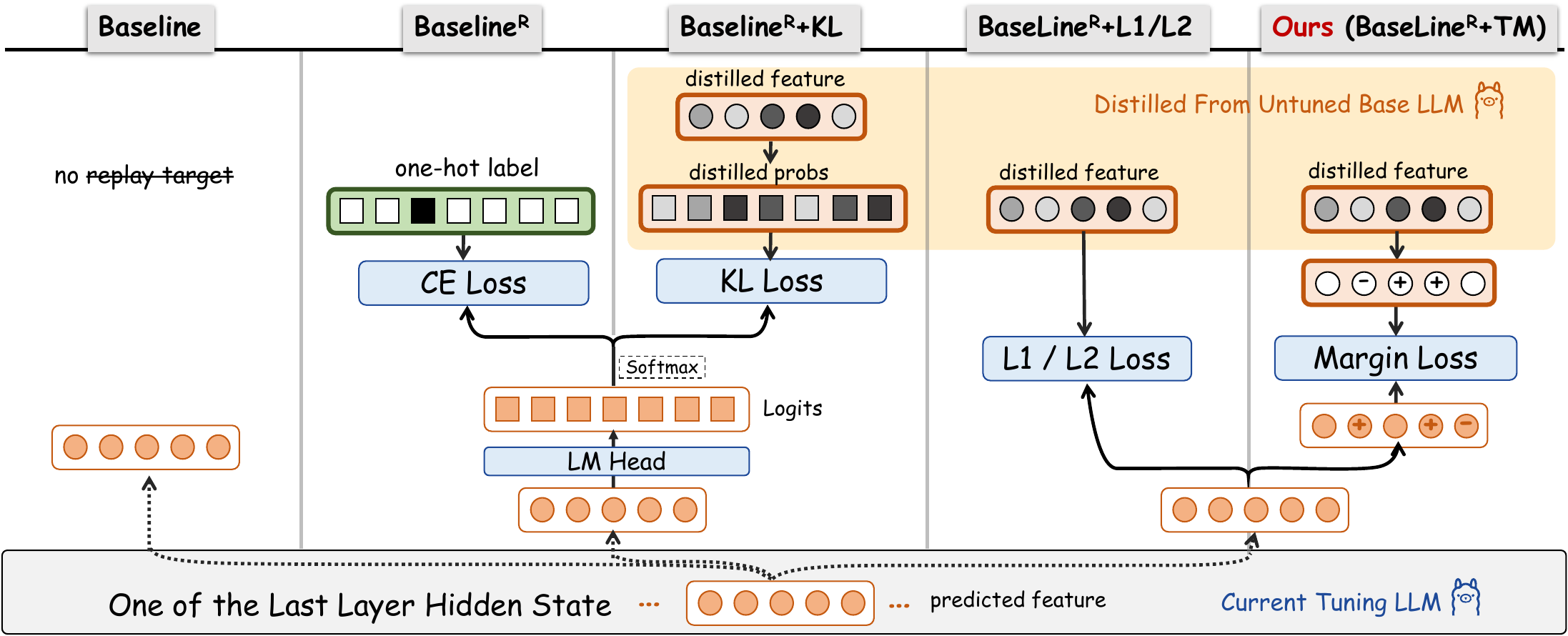}
    \caption{A comparable baseline series of distinct replay-based optimization targets (left to right): native non-replay Baseline, vanilla replay Baseline\textsuperscript{R}, replay with different distillation strategies regarding logits imitation Baseline\textsuperscript{R}+KL and feature imitation Baseline\textsuperscript{R}+L1/L2.
    The rightmost Baseline\textsuperscript{R}+TM is our proposed method, which employs the TM loss.
    }
    \label{fig:compare}
\end{figure*}

\subsection{Comparable Methods}
We meticulously implement all comparable methods from scratch for controlled experiments, including the most basic level and its progressively enhanced counterparts.
As shown in Fig.\ref{fig:compare}, we compare our method (denoted as Baseline\textsuperscript{R}+TM hereafter) with: native non-replay Baseline, vanilla replay Baseline\textsuperscript{R}, replay with different distillation strategies regarding logits imitation Baseline\textsuperscript{R}+KL and feature imitation Baseline\textsuperscript{R}+L1/L2.
These competitors cover the most prevalent and established practices in real-world application.
For fair comparison, all methods are implemented within the same framework using identical replay samples and maintaining consistent configuration throughout the evaluation process. Details are as follows:

\begin{itemize}[leftmargin=*, align=left]
    \item \textbf{Baseline}: continually finetune the LLM on sequential tasks without adding any general replay samples.
    \item \textbf{Baseline\textsuperscript{R}}: continually finetune the LLM on sequential tasks by mixing 1K general replay samples from $\mathcal{D}^{\text{(g)}}$ with each task. These samples are pre-selected before finetuning and remain unchanged throughout the entire finetuning process. Both the replay and downstream task samples are jointly optimized using the standard cross-entropy loss.
    Notably, all the methods discussed subsequently maintain this cross-entropy loss.
    
    \item \textbf{Baseline\textsuperscript{R}+KL}: extend the Baseline\textsuperscript{R} by integrating an additional KL loss.
    Specifically, the pre-distilled general replay sample logits serve as the target for KL loss during finetuning.
    The softmax temperature is set to 2, and the weight of KL loss term is accordingly set to 4 (its square) to compensate for gradient scaling down induced by the temperature~\cite{sun2024logit}.
    In implementation, to avoid the large overhead of pre-storing the high-dimensional $logits$ vectors, we compute the $logits$ in real-time during finetuning based on the previously acquired final layer hidden state $h^\text{(g)}$ and the original $lm\_head$ parameters of LLM.
    
    \item \textbf{Baseline\textsuperscript{R}+L1}: extend the Baseline\textsuperscript{R} by integrating an additional L1 loss computed on the hidden states at the last layer. The previously acquired $h^\text{(g)}$ serves as the target.
    \item \textbf{Baseline\textsuperscript{R}+L2}: resemble Baseline\textsuperscript{R}+L1 but employ L2 loss instead of L1 loss.
\end{itemize}
Our method and other options are explained as follows:
\begin{itemize}[leftmargin=*, align=left]

    \item \textbf{Our method (aka. Baseline\textsuperscript{R}+TM}): extend the Baseline\textsuperscript{R} by integrating our proposed TM loss computed on the hidden states at the last layer as in Eq.\ref{eq:tml}$\sim$Eq.\ref{eq:loss_final}.
    
    \item \textbf{BI Option}: adopt Batch Insertion (Sec.~\ref{sec:BI}) and evaluate its effectiveness across all the replay-based Baseline\textsuperscript{R} series, i.e., Baseline\textsuperscript{R} and Baseline\textsuperscript{R}+KL/L1/L2/TM that typically using general replay samples.

    \item \textbf{Loss Weight}: varied weighting values (denoted as w=[]) of the additional loss term regarding L1/L2/TM are tested. We first empirically set w=1 (omitted as default), w=100 and a dynamic weighting w=d.(Sec.~\ref{sec:DW}) for $\mathcal{L}^{TM}$ to find the optimal performance, and then deliberately evaluate the same optimal weight on L1 and L2 for fair comparison.

    \item \textbf{Upper Bound}: we also include the upper bound performance for comparison, where \textbf{Orig} denotes the original MMLU score of untuned base model as ceiling. \textbf{MTL} denotes a multi-tasks learning result across all 15 downstream tasks, which is trained on the combined task samples with the identical settings (epochs, learning rate, etc.). We calculate their F1 average upper bound as well.
\end{itemize}

\noindent Notably, Baseline\textsuperscript{R}+L1/L2 can be viewed as a stricter version of ours, which pursues precise value fitting (also bringing activation state alignment), but lacks the inherent variation tolerant afforded by our discrete states.

Regarding external competitors, since we strive for a simply yet effective replay-base approach (e.g., prompts-agnostic, task\_ids-agnostic, non-generative), we do not compare with ineligible methods like ProgPrompt~\cite{razdaibiedina2023progressive}, which sequentially integrates previously learned prompts with the current one during both training and testing.
Instead, we compare \textbf{O-LoRA}~\cite{wang2023orthogonal} in our LoRA setting due to its simplicity in only constraining the LoRA's update direction by an additional orthogonal loss term.
We are interested in comparing the downstream tasks performance enhanced as a byproduct by our method with that of the specialized O-LoRA, which is solely dedicated to this purpose.
We reimplement it carefully using the same LoRA hyperparameters.

\begin{table}[hp]
\footnotesize
\centering
\caption{Comparison of different methods on continual full-parameter finetuning (15 epochs per task) in 15 downstream tasks.}
\label{tab:continual_full}
\begin{tabular}{l|c|c|c}
\toprule
\makecell{Methods \\(Full-Parameter)}& \makecell{MMLU Score\\(Final)} & \makecell{15 Tasks AP\\(Final)} & \makecell{F1 Avg} \\
\midrule
\midrule
Baseline &
38.3213& 37.4720& 37.8919\\
\midrule

Baseline\textsuperscript{R} &
50.5332& 39.2741&44.1979\\
\rowcolor{gray!10}\quad w/ BI &
 55.5556&43.9903& 49.1011\\
\midrule

Baseline\textsuperscript{R}+KL &
 51.0492& 42.0231&46.0985\\
\rowcolor{gray!10}\quad w/ BI &
 52.7692 &35.5259&42.4638\\
\midrule

Baseline\textsuperscript{R}+L1 &
 54.9364&66.8605&60.3147\\
\rowcolor{gray!20}\quad w/ BI&
 54.5942& 66.7673&60.0691\\

Baseline\textsuperscript{R}+L2 &
55.0052 & 67.4899& 60.6113\\
\rowcolor{gray!10}\quad w/ BI &
 56.6219& 66.7462&61.2686\\

\midrule
Baseline\textsuperscript{R}+L1 (w=100)&
57.9635&72.4376&64.3973 \\
\rowcolor{gray!20}\quad w/ BI&
59.0299 &71.1125 & 64.5103 \\
Baseline\textsuperscript{R}+L2 (w=100) &
60.7499 &72.6112 &66.1531\\
\rowcolor{gray!20}\quad w/ BI&
57.8947 &73.2265 & 64.6643\\

\midrule
Baseline\textsuperscript{R}+L1 (w=d.)&
53.1132&67.4546&59.4309 \\
\rowcolor{gray!20}\quad w/ BI&
53.2852&64.4925&58.3556 \\
Baseline\textsuperscript{R}+L2 (w=d.) &
55.1772&71.0094&62.1001 \\
\rowcolor{gray!20}\quad w/ BI&
54.7988&68.2590&60.7927 \\

\midrule
\rowcolor{gray!40}
\textcolor{red!80!black}{\textbf{Ours}}& & &\\
Baseline\textsuperscript{R}+TM & 55.3836 &70.3490 & 61.9756 \\
\rowcolor{gray!10}\quad w/ BI & 57.6539& 68.7473&  62.7138\\

Baseline\textsuperscript{R}+TM (w=100) &
60.7155 & 74.0817 & 66.7359 \\
\rowcolor{gray!10}\quad w/ BI &
\textbf{60.9907}&72.4771&  66.2396\\

Baseline\textsuperscript{R}+TM (w=d.) &
60.7843 & \textbf{74.4796} & \textbf{66.9386} \\
\rowcolor{gray!10}\quad w/ BI &
57.2411 & 70.5607 & 63.2068 \\

\midrule
  
Upper Bound &
\makecell{Orig\\ 66.5291} & \makecell{MTL\\ 81.0079} & 73.0580\\

\bottomrule
\end{tabular}

\end{table}

\subsection{Implementation Details}
In all experimental comparison, we assess both full-parameter and LoRA finetuning settings with consistent batch size of 64.
All downstream samples are truncated with a maximum source length of 512 and a maximum target length of 50. Accordingly general replay samples are truncated with a maximum length of summation 562.
In full-parameter setting, we maintain a uniform learning rate of 3e-6 across all methods, employing a warmup strategy coupled with a cosine learning rate schedule.
In LoRA setting, we maintain a uniform learning rate of 1e-4 with warmup strategy across all methods, while seting LoRA hyperparameters to \textit{r}\,=8, \ensuremath{\alpha}=32, LoRA\_dropout=0.1 and only tuning the parameters limited to $q\_proj$ and $k\_proj$.
We aim to finetune each task with sufficient steps to ensure loss convergence. Based on preliminary experiments, we ultimately selected 15 epochs per task for the full-parameter setting (due to the smaller learning rate) and 8 epochs for the LoRA.

Notably, the replay samples used in all experiments are identical (the pre-selected set of 1K samples mentioned in Baseline\textsuperscript{R}), which guarantees that the observed effects are attributable to the methodological variations rather than differences in the replay data.
All experiments are conducted using Transformers~\cite{wolf2020transformers} library with DeepSpeed ZeRO-2~\cite{rajbhandari2020zero} and AdamW optimizer~\cite{loshchilov2017decoupled}, running on up to 8 H800-80GB GPUs.

For the BI option experiments, we set $\rho^{\text{BI}}$=$4/64$, which indicates that 4 general replay samples are inserted into each batch of 64 data points. Specifically, these 4 general replay samples are selected randomly and non-repetitively from the aforementioned 1K samples. Once all samples have been selected, the process is reset to ensure continuously sampling.

\subsection{Results}

\subsubsection{Continual Full-Parameter Finetune}
Table~\ref{tab:continual_full} shows the performance of each method in continual full finetuning 15 downstream tasks.
We find that simply mixing general replay samples (Baseline\textsuperscript{R}) significantly outperforms the Baseline without any replay, achieving a notable improvement of 12\% on MMLU. It justify the widespread adoption of this vanilla replay way in practice.

After additional distillation technique, Baseline\textsuperscript{R}+KL yields further improvements by capturing more information, specifically the distribution of labels.
However, under the similar distillation cost, feature-based methods (Baseline\textsuperscript{R}+L1/L2/TM) perform remarkable better,
empirically suggesting that feature information is more efficient than label information by encoding richer representations of model knowledge in feature layer.
Our rationale is that softmax-normalized labels tend to be dominated by extreme values, whereas features preserve finer-grained details.
Among them, Baseline\textsuperscript{R}+TM further alleviate the overly rigid inherence of L1/L2 loss by appropriately constraining optimization from an activation state perspective, achieving the highest performance.

Moreover, the results of all Baseline\textsuperscript{R} series validate the hypothesis that using only general replay samples can simultaneously maintain general capabilities and enhance overall performance of downstream task. As shown in the table, both MMLU and AP improve.
This provides an alternative to the traditional practice of laboriously collecting downstream task replay samples during continual finetuning, sparking a promising research direction.

\begin{table}[t]
\footnotesize
\centering
\caption{Comparison of different methods on continual LoRA finetuning (8 epochs per task) in 15 downstream tasks (w=d. denotes weight dynamic).}
\label{tab:continual_lora}
\begin{tabular}{l|c|c|c}
\toprule
\makecell{Methods \\(LoRA)}& \makecell{MMLU Score\\(Final)} & \makecell{15 Tasks AP\\(Final)} & \makecell{F1 Avg} \\

\midrule
\midrule
Baseline &
55.7620 & 73.3944 & 63.3746\\
\midrule
Baseline\textsuperscript{R} &
58.6515& \textbf{75.5310}& 66.0296\\

\rowcolor{gray!10}\quad w/ BI &
56.5187& 73.3986& 63.8621\\
\midrule

Baseline\textsuperscript{R}+KL &
61.0251& 74.8367& 67.2289\\

\rowcolor{gray!10}\quad w/ BI &
61.5755& 72.9626& 66.7872\\

\midrule
Baseline\textsuperscript{R}+L1 &
61.5411& 73.4170& 66.9565\\

\rowcolor{gray!10}\quad w/ BI &
65.1875& 73.1178& 68.9253\\

Baseline\textsuperscript{R}+L2 &
61.6787& 74.9397& 67.6656\\

\rowcolor{gray!10}\quad w/ BI &
64.4651& 74.1000& 68.9476\\

\midrule
\rowcolor{gray!40}
\textcolor{red!80!black}{\textbf{Ours}} & & &\\
Baseline\textsuperscript{R}+TM &
65.3251 & \textbf{75.0639}& \textbf{69.8567} \\
\rowcolor{gray!10}\quad w/ BI &
64.6371 & 72.7650 & 68.4606\\

Baseline\textsuperscript{R}+TM (w=100) &
65.9443& 63.9167& 64.9147 \\
\rowcolor{gray!10}\quad w/ BI &
65.5659& 68.7755& 67.1323\\

Baseline\textsuperscript{R}+TM (w=d.) &
\textbf{66.2539}& 64.4417&65.3352 \\
\rowcolor{gray!10}\quad w/ BI &
65.4627& 67.7580& 66.5906\\

\midrule
O-LoRA~\cite{wang2023orthogonal} &
55.8996 & 73.6823 & 63.5707\\

\midrule
Upper Bound &
\makecell{Orig\\ 66.5291} & \makecell{MTL\\ 80.3474} & 72.7882\\

\bottomrule
\end{tabular}

\end{table}

\subsubsection{Continual LoRA Finetune}
Table~\ref{tab:continual_lora} shows the performance of each method in continual LoRA finetuning 15 downstream tasks.
Compared to full-parameter setting, LoRA alone exhibits notable superiority, which tunes only 0.042\% of the parameters (i.e., $q\_proj$ and $k\_proj$), This minimal parameter tuning likely contributes to its strong anti-forgetting ability, but it still enables adequate learning of new tasks.
For instance, when equipped with LoRA, both the Baseline and vanilla replay Baseline\textsuperscript{R} nearly match the best F1 Avg observed in full-parameter, and Baseline\textsuperscript{R} also show surprisingly strong AP of downstream tasks.
Still, similar trends hold for LoRA, with the Baseline\textsuperscript{R} series showing progressive improvements, where our method ultimately achieves the best F1 Avg.
We also find that several methods here have achieved MMLU scores nearly matching the upper bound evaluated from original base model, showing negligible loss of general capabilities.

O-LoRA, as a simple and comparable approach dedicated to AP of downstream tasks, achieves decent AP performance but exhibits obvious forgetting of general capabilities. Furthermore, the original O-LoRA paper claims its superiority over the method with replaying downstream task samples, but we still attain a higher AP. This demonstrates the multifaceted advantages of our GeRe framework over tradition.

By the way, the MTL performance under both settings shows that LoRA still slightly underperforms full-parameter when jointly learning multiple new tasks, aligning with study\cite{biderman2024lora} and suggesting that full-parameter remains preferable in normal situation with available computational resources.

\subsubsection{Ablation Study of BI and Loss Weight}
Each method with BI option under full-parameter and LoRA settings is additional list in Table.\ref{tab:continual_full}$\sim$\ref{tab:continual_lora}.
The effectiveness of BI varies across methods and settings, without showing consistent enhancement. For instance, BI improves the Baseline in full-parameters and Baseline\textsuperscript{R}+L1/L2 in LoRA, but its effect appears negligible in most distillation methods that already capture more information.
We attribute this to the small scale of our finetuning datasets in the experiments, where mixing 1K general replay samples suffices for many downstream tasks. In extremely small datasets like CB, BI’s proportional insertion may even reduce the final replay samples below the standard 1K.

However, BI remain necessary in potential scenarios especially finetuning large-scale downstream task datasets that may far exceeds the 1K replay samples. In such cases, data balancing is crucial since simply mixing them at their original scale leads to insufficient replay. 
As shown in the results, while BI does not significantly improve performance, it also does not degrade it especially with our method, indicating that Baseline\textsuperscript{R}+TM with BI can be directly used in most circumstances.
In conclusion, the adoption of BI should be determined by practical considerations and an optimal replay insertion ratio, which warrants further investigation

Regarding different loss weight, the results also vary.
In full-parameter setting, our method with dynamic weighting $\mathcal{L}^{TM}$, i.e., Baseline\textsuperscript{R}+TM (w=d.) performs best, even in a fair comparison where L1 and L2 are purposefully evaluated with the same weighting strategy.
In contrast, a simply setting of fixed weight of 1 yields better performance under LoRA, as larger or dynamic weight tend to degrade AP. So We only list results (w=1) of L1 and L2 as well. This indicates that different settings should better have their individual weight strategies, and we have already explore two best practice. Unified settings remain for future research. In the following comparison, we intentionally use results of w=100 for full-parameter setting and w=1 for LoRA in order to simultaneously consider the maximum achievable performance of the L1/L2 competitors.

Furthermore, conventional belief suggests that strengthening the weight of optimization direction toward anti-forgetting will enhance stability at the cost of plasticity, thereby impairing learning of new task. However, results with higher weight (from w=1 to w=100) show that not only MMLU dose but also downstream task's AP continues to improve. This may stem from the adoption of general replay samples rather than task-specific replay samples, which mitigates the Stability-Plasticity Dilemma \cite{dohare2024loss} typically occurs when excessively replaying samples from downstream tasks in tradition.

\begin{figure*}[htbp]
    \centering
    \includegraphics[width=\textwidth]{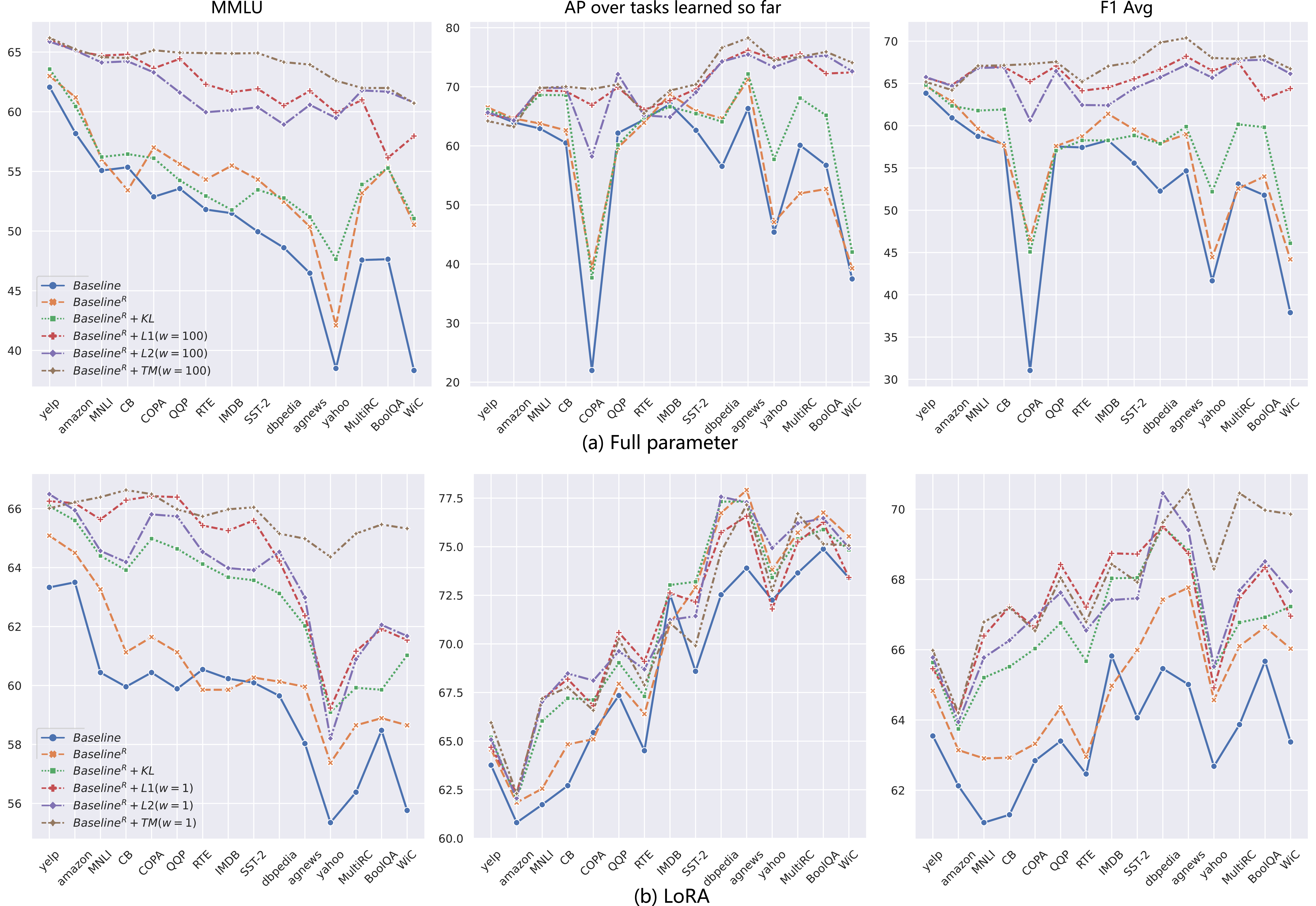}
    \caption{Performance trend during continual learning 15 tasks of different methods. Y-axis of each figure indicates the specific task that has just been learned. Two rows depict full-parameter and LoRA settings, respectively. Three columns are metrics: current MMLU score, average performance over tasks learned so far, F1 average.}
    \label{fig:over_tasks}
    \vspace{-10pt}
\end{figure*}

\begin{figure}[htbp]
    \centering
    \includegraphics[width=0.9\columnwidth, height=8.5cm]{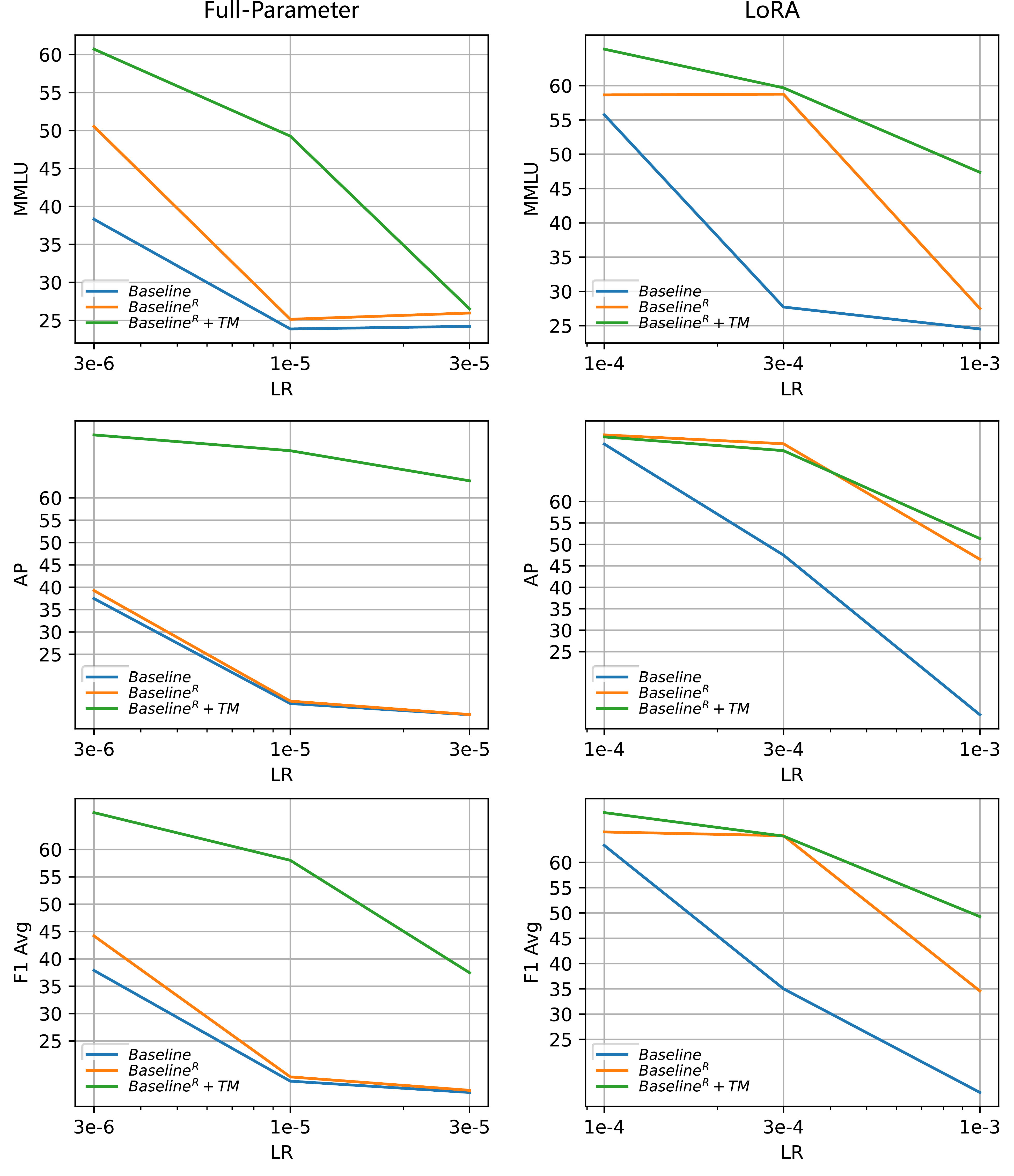}
    \vspace{-8pt}
    \caption{MMLU, AP and F1 Avg performance of three major representative methods across different learning rate is compared under full-parameter and LoRA settings, with the LR axis displayed on a logarithmic scale.}
    \label{fig:lr_robustness}
    \vspace{-10pt}
\end{figure}

\begin{figure*}[ht]
    \centering
    \includegraphics[width=1\textwidth]{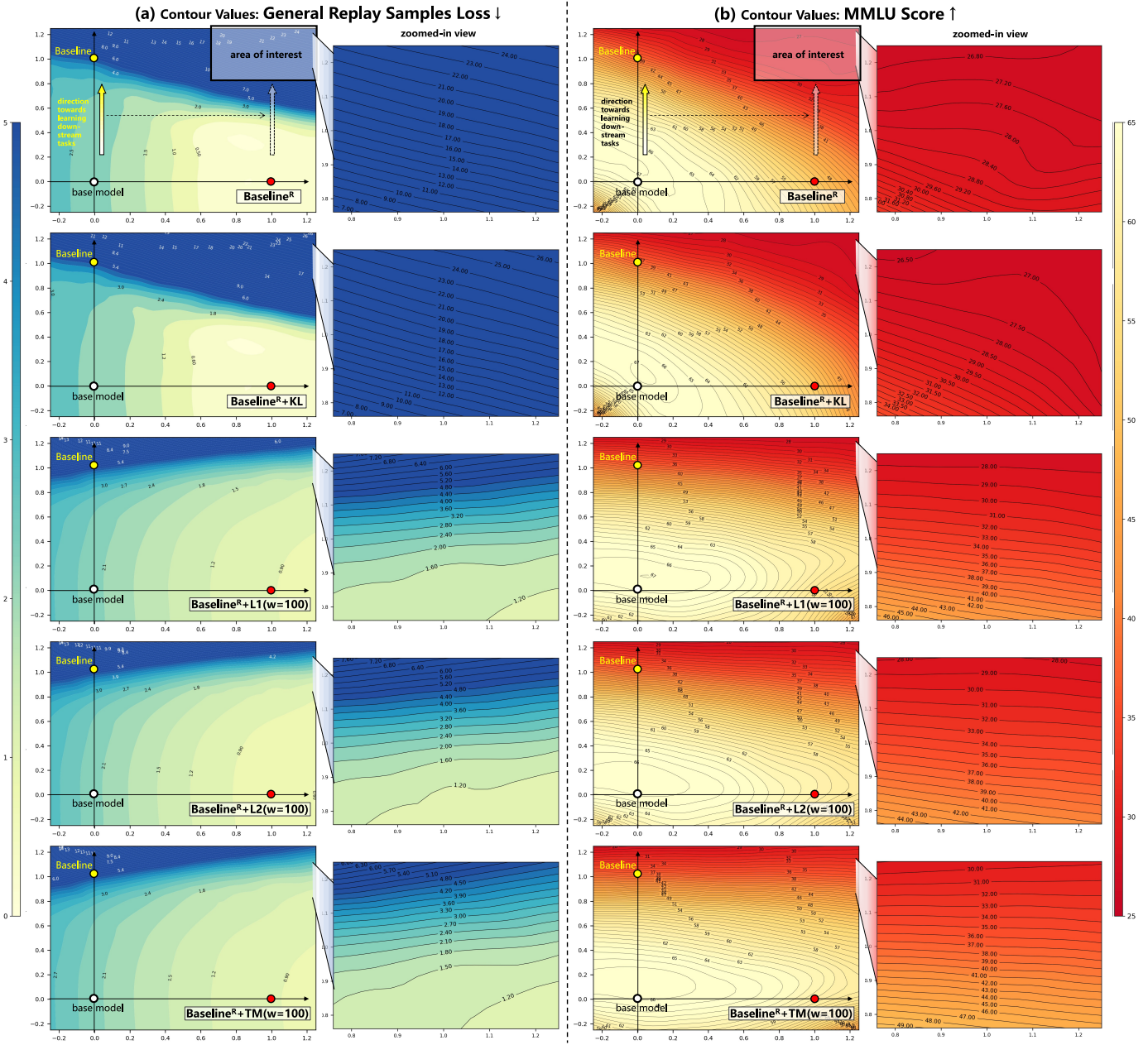}
    \caption{Landscapes of (a) replay samples loss, and (b) MMLU score under \textbf{full-parameter setting}.
    Origin point (0,0) is base untuned model.
    Y-axis is weight update direction of Baseline (0,1), representing the learning dedicated to downstream tasks.
    X-axis is weight update direction of target method for comparison (1,0).
    The upper-right area of interest simulates the target model guided by the learning direction of downstream tasks (yellow arrow), where the flatness (see zoomed-in view) can imply the optimizing robustness against latent forgetting even under potential overtraining in practice.
    }
    \label{fig:landscape_full}
\end{figure*}

\begin{figure*}[ht]
    \centering
    \includegraphics[width=1\textwidth]{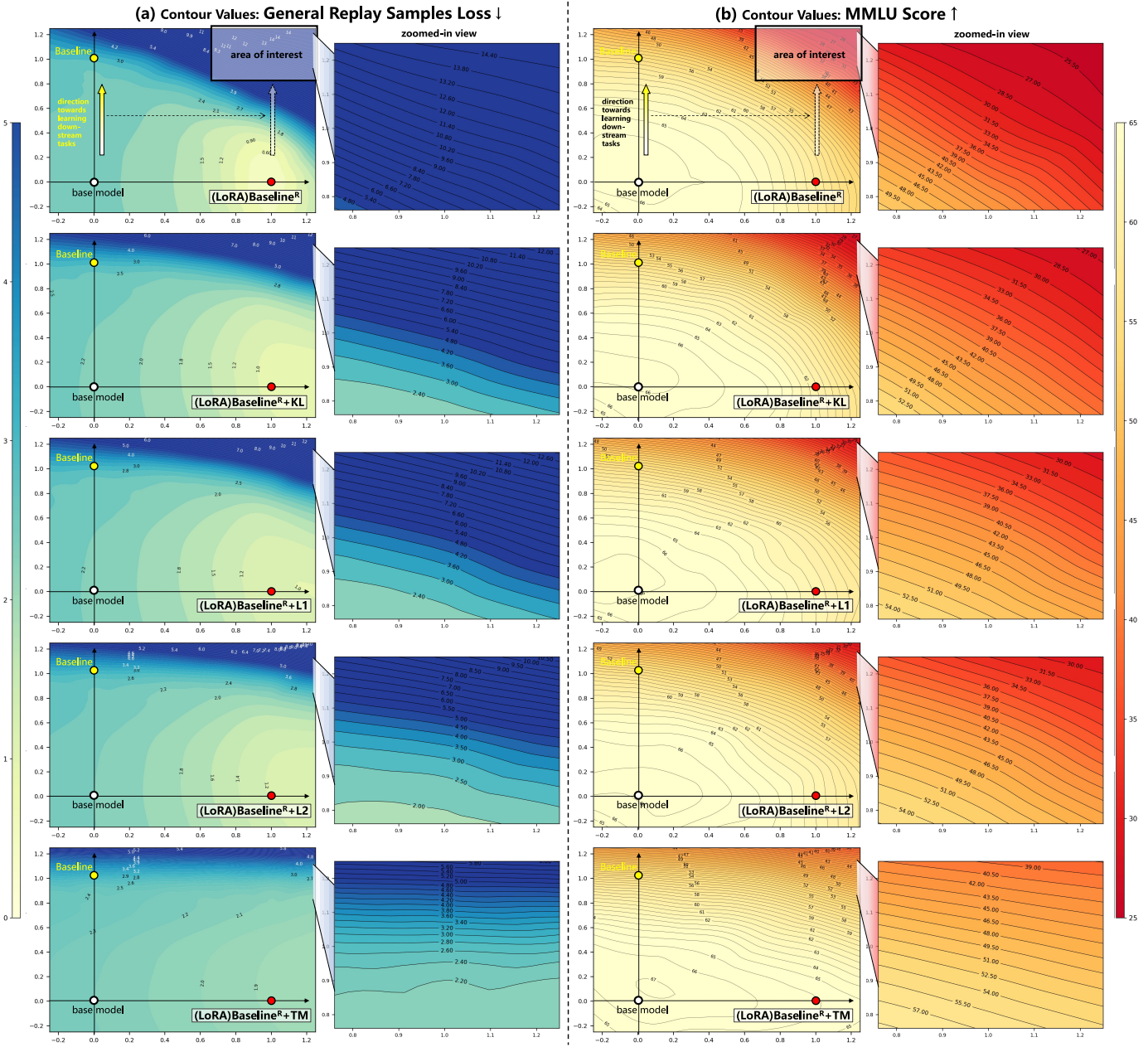}
    \caption{Landscapes of (a) replay samples loss, and (b) MMLU score under \textbf{LoRA setting}.
    Origin point (0,0) is base untuned model.
    Y-axis is weight update direction of Baseline (0,1), representing the learning dedicated to downstream tasks.
    X-axis is weight update direction of target method for comparison (1,0).
    The upper-right area of interest simulates the target model guided by the learning direction of downstream tasks (yellow arrow), where the flatness (see zoomed-in view) can imply the optimizing robustness against latent forgetting even under potential overtraining in practice.
    }
    \label{fig:landscape_lora}
\end{figure*}

\subsubsection{Performance Trend Over Tasks}
Fig.\ref{fig:over_tasks} shows the dynamic changes of the three metrics assessed in Table \ref{tab:continual_full}$\sim$\ref{tab:continual_lora} as the model sequentially learns 15 downstream tasks under full-parameter and LoRA settings.
Evidently, our method consistently achieves the highest score on MMLU at nearly every task step under both settings. Moreover, it remains in the top tier performance of AP across the downstream tasks learned so far, achieving the best final F1 Avg.

Notably, during the full-parameter learning, a significant decline in AP occurs after the model learned the COPA task.
A Deeper investigation of the task-wise results (see Appendix) reveals that it is mainly caused by performance drops in the MNLI and CB tasks.
We attribute this to the unique instruction format of COPA without providing options (see Appendix), which temporarily disrupts the model's instruction-following ability after learning COPA.
So, tasks with the most similar instructions like MNLI and CB experience performance degradation.
Fortunately, as the model continues to learn subsequent tasks with regular instructions, the performance of these two tasks recovers. We interpret this as a case of spurious forgetting~\cite{zheng2025spurious}, where the model does not lose the core knowledge of these tasks but undergoes temporary confusion in instruction following, which can be readily restored in later learning phases.

\subsubsection{Robustness to Learning Rate}
In continual finetuning, it is well-established that while larger learning rates (LR) facilitate more thorough learning of downstream tasks, they also intensify the forgetting of previously acquired knowledge.
This effect becomes particularly pronounced when dealing with LLMs featuring massive training parameters, which aligns with our observations in preliminary experiments.
Thus, practitioners need to carefully adjust the LR from relatively small values to balance new task acquisition with knowledge retention. 

However, empirical evaluating against both the native Baseline and vanilla replay Baseline\textsuperscript{R} show that our method maintains relatively strong general capabilities (MMLU scores) even with substantially increased LRs.
As shown in Fig.\ref{fig:lr_robustness}, our method gains more stable performance despite a 3× LR increase under full-parameter and a 10× increase under LoRA.
In contrast, the compared methods approach a MMLU score of nearly 25\% ,equivalent to random guessing among 4 options, highlighting their vulnerable dependence on tuning LR.
Our methods demonstrates superior adaptability in practice scenarios.
Beyond MMLU, similar conclusions regarding AP and the resulting F1 Avg can be drawn from the subsequent subfigures.

\subsubsection{Robustness in Optimization Landscape}
To better understand the underlying optimization mechanisms of different methods and their robustness against forgetting, we visualize the landscapes with two contour values under full-parameter (Fig.\ref{fig:landscape_full}) and LoRA settings (Fig.\ref{fig:landscape_lora})

Our idea is that in replay-based learning, to ensure thorough learning for downstream tasks, excessive training can easily occur, which will compromise the general capabilities.
We consider that a better method should reconcile the optimization directions of both downstream tasks and replay samples. Such a method would maintain latent robustness even when subjected to excessive downstream task training that typically induces forgetting.
According to task vector arithmetic~\cite{ilharco2023editing}, we can directly perform linear combinations of model weight for a specific method to simulate and observe their robustness to optimization dynamics under extreme conditions of excessive training. 
Therefore, the landscape is designed as a 2D weight space spanned by two specific model weight update directions, where y-axis is the update direction of the Baseline and x-axis is one of the interested methods for comparison.
Specifically, the upward direction (yellow arrow) indicates the optimization toward native continual finetuning without any replay, highlighting exclusive learning of downstream tasks.
The rightward direction indicates the optimization toward a target model trained from a specific method among the replay-based series.
Based on this coordinate, the upper-right region (white rectangle) is the area of interest that can reveal the robustness against forgetting undergoing potential overtraining, as it simulates the weight update direction imposed on the target model toward overly optimizing downstream tasks.

As for contour values, we select two metrics: a) the CE loss of replay samples, as it is universally adopted and optimized across all methods, and b) the direct MMLU score for straightforward observation. These values measure the retention of general capabilities implicitly and explicitly, respectively. The flatness (i.e., the rate of performance degradation) of the area of interest indicates how robustly each method preserves their general capabilities while learning downstream task.

The landscapes are implemented with positioning the untuned base LLM model at coordinate (0,0), the native Baseline finetuned model at coordinate (1,0), and a specific replay-based finetuned model for comparison at coordinate (0,1).
Weight parameters of each model are flattened into a vector, and the two basis vectors of this coordinate system are derived by subtracting the corresponding model weight vectors
(e.g., $\vec{\mathbf{y}}{=}\mathbf{w}_{\text{[target]}}{-}\mathbf{w}_{\text{base}}, \vec{\mathbf{x}}{=}\mathbf{w}_{\text{baseline}}{-}\mathbf{w}_{\text{base}}$).
We use these two basis vectors to generate a grid of points, where each point represents a model derived from a linear combination of these basis vectors
(e.g., (0.6,0.4) denotes model of weight $\mathbf{w}{=}0.6\vec{\mathbf{x}}{+}0.4\vec{\mathbf{y}}$).
For every model associated with the points, we compute the loss of replay samples and the MMLU score, creating the contour plot as landscape, respectively.

For instance, Fig.\ref{fig:landscape_full}(a) shows the replay sample loss under full-parameter setting.
We observe that the Baseline\textsuperscript{R} exhibits a notably steepness in the area of interest, implying that this optimization encounters significant conflicts when following the direction of learning downstream tasks while attempting to replay to retain general capabilities. This potentially accounts for its poor performance in Tab.\ref{tab:continual_full}.
The same issue persists in Baseline\textsuperscript{R}+KL, though it is relatively less severe, but still remarkable.
In comparison, the feature-based replay methods show significantly flatter behavior in the area of interest, and among them our Baseline\textsuperscript{R}+TM performs the flattest upon closer look at the contour values in the zoomed-in view. This mean our method show better robustness to the intrinsic optimization dynamics when arbitrarily or excessively trained on downstream tasks.

For the MMLU score landscape in Fig.\ref{fig:landscape_full}(b), similar trend is observed. The feature-based replay methods exhibit higher scores and slower decline in the area of interest.
They also shape a distinct ridge along the learning trajectory (i.e., y-axis) of the specific model, where the scores are maximally preserved.
Besides, though the learning trajectories of baseline\textsuperscript{R} and baseline\textsuperscript{R}+KL maintain a high score early, they undergo sharply decline as more tasks are introduced. In contrast, the feature-based methods effectively preserve the score along the ridge, confirming the necessity of benchmarking typical long-sequence tasks.
In the MMLU score landscape, our Baseline\textsuperscript{R}+TM still demonstrate superior performance.

Additionally, although the contour patterns of general samples loss and MMLU score differ, their underlying trends exhibit similar characteristics, e.g., both metrics show consistent variations in the area of interest of the same method. This confirms that general samples can implicitly reflect the actual general capabilities.
However, relying solely on the CE loss of general samples may be insufficient, e.g., Baseline\textsuperscript{R} and Baseline\textsuperscript{R}+KL achieve lower loss values, but their MMLU scores remain low.
Instead, the optimization of feature-based methods align more closely with the trends of MMLU.

Finally, Fig.\ref{fig:landscape_lora} shows landscape under LoRA setting, where the observations are generally similar to those of full-parameter, except that LoRA---as a highly effective anti-forgetting tool---significantly enhances the foundational performance of all variants. Notably, our method here shows more pronounced flatness and maintains higher performance in the are of interest. Across both settings, our Baseline\textsuperscript{R}+TM consistently demonstrates the optimizing robustness against latent forgetting.

\section{Conclusion}\label{sec:Conclusion}
In this research, we introduce GeRe, a framework that leverages general replay samples for continual learning in LLMs.
Building upon GeRe, we revisit the existing replay baseline and devise a novel optimization method that utilizes the informative states of neurons through a proposed TM loss.
This loss function effectively aligns the activation states of replay samples, offering a moderate yet discerning constraint compared to existing replay-based variants.
Crucially, GeRe’s results reveal that only a fixed set of general replay samples is sufficient for continual learning across a long sequence of downstream tasks, which not only effectively retains the general capabilities but also successfully promotes the overall performance on downstream tasks.
Furthermore, detailed analyses and intuitive visualizations rigorously validate the superior performance and robustness of the TM loss within GeRe.
Our study offers valuable insights into the efficacy of replay mechanisms, highlighting the practical advantages and contributing to potential applications for the continuous iteration of LLMs.

\ifCLASSOPTIONcaptionsoff
  \newpage
\fi

\bibliographystyle{IEEEtran}
\bibliography{main}

\appendix

\section*{A}\label{sec:appendix}

\subsection{Examples of General Replay Samples}
Table.\ref{tab:detailed_gen_samples} presents some selected examples from the general replay samples set $\mathcal{D}^{\text{(g)}}$ used throughout this paper, where each entry is a normal pretraining text sentence. The ID is the line number of our released \textit{jsonl} file of $\mathcal{D}^{\text{(g)}}$, and the set name is the meta information indicating the source.
(We purposely selected examples from diverse sources for display.)
The data was obtained through the following process: We downloaded the first chunk of data (\textit{train-00000-of-00048-ab2b35705f029d94.parquet}) from SlimPajama-6B (\url{https://huggingface.co/datasets/DKYoon/SlimPajama-6B}), a sampled version of SlimPajama-627B (\url{https://huggingface.co/datasets/cerebras/SlimPajama-627B}). Then, we simply extracted the first 1000 entries using the following code to generate the \textit{jsonl} file:
\begin{lstlisting}
slim_datasets = load_dataset('parquet', 
    data_files={'train-00000-of-00048-ab2b35705f029d94.parquet'}
)['train']
slim_datasets.select(range(1000)).to_json('slimpajama_6B_chunk0_head1k.jsonl')
\end{lstlisting}
This acquisition process demonstrates that the general replay sample set was obtained through random selection rather than deliberate curation, thereby substantiating the robustness and universality of our method regarding the replay data.
\newcommand{\n}{\textbackslash n}
\begin{table*}[htbp]

\centering

\resizebox{\linewidth}{!}{

\begin{tabular}{c|c|c}
\toprule
Id & Set Name & Text\\
\midrule
\midrule
0 & RedPajamaC4 & \parbox{15cm}{Want Tori to Coach You Too?\n Tori's Health Step by Step coming soon.\n Win free copies, prizes, access to exclusive behind-the-scenes, free access to Coach Tori, and more.\n and receive a copy of Tori's Weekly Challenges. We'll also notify you of when Tori's Program becomes available.\n I've been asked, even criticized, about adding a focus on nutrition to Desert. There's a reason why. I had poor nutritional examples growing up. Being confused on the issue of nutrition cost me a lot. I remember yo-yo'ing a lot. The only time I even came close to being my desired weight was when I did high-intensity workouts daily. At one point, I was exercised about 6 hours a day. I was in multiple dance classes and a karate class, as well as another karate club that met for two hours three days a week. I also rode my bike to campus, and even added a one hour workout when I got home. I was still thirty pounds overweight. I can attest to the coined phrase "You cannot exercise away a bad diet."\n It was hard to consider diet for me, because I had a genetic heritage that leaned on the heavy side. I felt trapped, having a low metabolism. It seemed if I even looked at what others ate, I was the one who gained weight.\n Every once in a while, someone would mention diet to me, but it did little to sway me. Why? Bad examples.\n Brad Pitt in Ocean's Eleven. Every scene he's in, he's eating something unhealthy.\n In Hollywood and at school, lots of "lean" people were eating the things I loved: pizza, ice cream, hamburgers, fries, bread, cake, cookies, etc.\n I also knew several "weighed down" people who were eating a healthy diet.\n It wasn't until I was in college, having just finished my laps in swimming plus a jog, that I stopped by to visit a friend--a slim friend who never seemed hungry. It seemed so unfair as I watched her prepare herself a salad and two small slices of pizza.\n I knew in that moment that if I ate like her, I would look like her. I also knew that if I prepared a small salad and two small slices of pizza, that by the end of the meal I would end up not eating just one personal pizza, but two or three.\n I started to believe in nutrition, but I didn't have faith that someone like me could do it.\n I was right, and I was wrong . . .}\\
\midrule
25 & RedPajamaGithub & \parbox{15cm}{ACCEPTED\n \n \#\#\#\# According to\n The Catalogue of Life, 3rd January 2011\n \n \#\#\#\# Published in\n New Zealand J. Bot. 25:166. 1987\n \n \#\#\#\# Original name\n Atropis pumila Kirk\n \n \#\#\# Remarks\n null}\\
\midrule
500 & RedPajamaC4 & \parbox{15cm}{Located in an impressive old draper's warehouse, Citibase Birmingham Mailbox is in the heart of the vibrant Mailbox shopping, entertainment and dining district. The recently refurbished reception area is reminiscent of a New York warehouse and the centre provides a wide range of offices many with amazing city views, including new Loft-style suites.\n With New Street Station and the smart new Grand Central Shopping Centre, the Central Business District and the vast array of other shopping and dining options all under 10 minutes' walk away, it's the ideal location to grow your business.}\\
\midrule
839 & \shortstack[l]{RedPajama\\CommonCrawl} & \parbox{15cm}{Big Boy's 24/7 Channel\n Real 92.3 LA\n BigBoyTV Videos\n Big Boy's Bankroll\n Big Boy Full Episodes\n Big Boy's Fully Loaded Interviews\n Big Boy's Uncut Podcast\n What's Trending with Natalia Perez\n Meet the Neighborhood\n Natalia Perez\n Vick One\n DJ Hed\n Advertise on Big Boy's Neighborhood\n Podcast: Home Grown Radio\n Tupac's "Strictly 4 My NIGGAZ" Will Be Out Again For Its 25th Anniversary\n By DJ Hed Feb 17, 2018\n Yesterday (February 16) marked the 25th anniversary of Tupac's sophomore album, Strictly 4 My N.I.G.G.A.Z.\n Today, Interscope Records and UMe are gifting Pac and hip-hop fans are around the world, the blessing to cop a limited edition, commemorative 2LP vinyl of the project.\n There's two vinyl editions available for purchase. The standard edition is available at all physical retailers and comes with the 180-gram vinyl of the album, and the deluxe edition features a gatefold image of 2PAC's original notebook, with his handwritten track list visible, and prints.\n The only catch is that there's only 1,000 copies made, which means you have to cop your copy asap. Grab your Strictly 4 My N.I.G.G.A.Z copy on Tupac's website here.\n About DJ Hed\n DJ Hed is a deejay mixer on REAL 92.3 KRRL FM Los Angeles RadioRead More\n Big Boy Blog\n Big Boy's Full Episodes\n BIGBOY Political File\n \textbackslash u00a9 2021 Premiere Networks, Inc.}\\
\midrule
1000 & RedPajamaC4 & \parbox{15cm}{The Pastel Piebald is a co-dom recessive morph combination, we produced it in 2005 along with The Snake Keeper. After missing the odds on multiple clutches, our luck changed with the second to last clutch of the season, from a 5 egg clutch of Pastel het Pied x het Pied, out came one of our prized possessions one of the first Pastel Pieds. You can imagine the excitement and joy that was felt on that Labor Day holiday in 2005, when we discovered this beautiful Pastel Pied had hatched and it was a male. This male has grown up and in 2008 had sired the first Super Pastel Pied or "Killer Pied", a stunning lemon yellow Piebald, creating a greater demand for the already sought after Pastel Pied.}\\
\bottomrule
\end{tabular}

}
\caption{The example of the adopted general replay samples}
\label{tab:detailed_gen_samples}
\end{table*}

\subsection{Examples of Downstream Task Datasets}
Table.\ref{tab:detailed_tasks1}\textasciitilde\ref{tab:detailed_tasks2} show detailed examples of 15 downstream task datasets, including task types, dataset names, instructions, inputs, and golden answers.
All data are constructed using the same template:

\scalebox{0.9}{\texttt{[Instruction]\textbackslash n[Input]\textbackslash nAnswer:[Golden Answer]}.}

\noindent The evaluation criterion for all samples is binary classification for being correct or incorrect , determining whether the model-generated answers exactly match the golden answers. The accuracy for each dataset is then calculated as the corresponding task performance.

\begin{table*}[htbp]
\centering

\resizebox{\linewidth}{!}{

\begin{tabular}{c|c|c|c|c}
\toprule
Task Type& Dataset & Instruction & Input & Golden Answer\\
\midrule
\midrule
SC & yelp & \parbox{5cm}{What is the sentiment of the following paragraph? Choose one from the option.\\Option: very negative, negative, neutral, positive, very positive} & \parbox{5cm}{Text: This place is printing money and rightfully so. They simply do a bang up job. Best BBQ in AZ.} & very positive \\
\midrule
SC & amazon & \parbox{5cm}{What is the sentiment of the following paragraph? Choose one from the option.\\Option: very negative, negative, neutral, positive, very positive} & \parbox{5cm}{Title: Very fragile...arrived broken\\Text: The set is cute, but refrigerator door was broken on arrival and not repairable. The table top and hutch had come apart and required regluing. This set will not stand up to play.} & negative\\
\midrule
NLI & MNLI & \parbox{5cm}{What is the logical relationship between the ``sentence 1'' and the ``sentence 2''? Choose one from the option.\\Option: neutral, entailment, contradiction} & \parbox{5cm}{sentence 1: She leaned back in her chair.\\sentence 2: She stood next to a chair.} & neutral \\
\midrule
NLI & CB & \parbox{5cm}{What is the logical relationship between the ``sentence 1'' and the ``sentence 2''? Choose one from the option.\\Option: entailment, contradiction, neutral} & \parbox{5cm}{sentence 1: A: Your turn. B: Okay. Uh, I don't think they should abolish it.\\sentence 2: they should abolish it} & contradiction \\
\midrule
COPA & COPA & \parbox{5cm}{} & \parbox{5cm}{Which sentence is the cause of ``I coughed.''? Choose one between A and B.\\A: I inhaled smoke.\\B: I lowered my voice.} & A \\
\midrule
QQP & QQP & \parbox{5cm}{Whether the ``first sentence'' and the ``second sentence'' have the same meaning? Choose one from the option.\\Option: False, True} & \parbox{5cm}{first sentence: What are the best franchises in India?\\second sentence: What are the best franchise in India?} & True \\
\midrule
NLI & RTE & \parbox{5cm}{What is the logical relationship between the ``sentence 1'' and the ``sentence 2''? Choose one from the option.\\Option: contradiction, entailment} & \parbox{5cm}{sentence 1: The girl was found in Drummondville.\\sentence 2: Drummondville contains the girl.} & contradiction \\
\midrule
SC & IMDB & \parbox{5cm}{What is the sentiment of the following paragraph? Choose one from the option.\\Option: Good, Bad} & \parbox{5cm}{This is a good film. This is very funny. Yet after this film there were no good Ernest films!} & Good \\
\midrule
SC & SST-2 & \parbox{5cm}{What is the sentiment of the following paragraph? Choose one from the option.\\Option: Good, Bad} & \parbox{5cm}{Text: it 's not the ultimate depression-era gangster movie .} & Bad \\
\midrule
TC & dbpedia & \parbox{5cm}{What is the topic of the following paragraph? Choose one from the option.\\Option: Company, Educational Institution, Artist, Athlete, Office Holder, Mean of Transportation, Building, Natural Place, Village, Animal, Plant, Album, Film, Written Work} & \parbox{5cm}{Title: Cori Schumacher\\Text:  Cori Schumacher is a world champion surfer from California.} & Athlete \\
\midrule
TC & agnews & \parbox{5cm}{What is the topic of the following paragraph? Choose one from the option.\\Option: World, Sports, Business, Science or Technology} & \parbox{5cm}{Title: British sailors bag bronze\\Text: Britain's Chris Draper and Simon Hiscocks win bronze in a tense final 49er race on the Saronic Gulf.} & Sports \\
\midrule
TC & yahoo & \parbox{5cm}{What is the topic of the following paragraph? Choose one from the option.\\Option: Society \& Culture, Science \& Mathematics, Health, Education \& Reference, Computers \& Internet, Sports, Business \& Finance, Entertainment \& Music, Family \& Relationships, Politics \& Government} & \parbox{5cm}{Title: did God create people or did people create god?..?\\Question: think about it...\\Answer: Good question dude.} & Society \& Culture \\
\bottomrule
\end{tabular}

}
\caption{The example of the 15 downstream tasks for fine-tuning}
\label{tab:detailed_tasks1}
\end{table*}

\begin{table*}[htbp]
\centering
\resizebox{\linewidth}{!}{

\begin{tabular}{c|c|c|c|c}
\toprule
MultiRC & MultiRC & \parbox{5cm}{According to the following passage and question, is the candidate answer true or false? Choose one from the option.\\Option: False, True} & \parbox{5cm}{paragraph: Susan wanted to have a birthday party. She called all of her friends. She has five friends. Her mom said that Susan can invite them all to the party. Her first friend could not go to the party because she was sick. Her second friend was going out of town. Her third friend was not so sure if her parents would let her. The fourth friend said maybe. The fifth friend could go to the party for sure. Susan was a little sad. On the day of the party, all five friends showed up. Each friend had a present for Susan. Susan was happy and sent each friend a thank you card the next week. \\question: Did Susan call her friends before or after asking her mother?\\candidate answer: Before asking her mother} & True \\
\midrule
BoolQA & BoolQA & \parbox{5cm}{According to the following passage, is the question true or false? Choose one from the option.\\Option: True, False} & \parbox{5cm}{question: can u drive in canada with us license\\passage: American entry into Canada by land -- Persons driving into Canada must have their vehicle's registration document and proof of insurance.} & True \\
\midrule
WiC & WiC & \parbox{5cm}{Given a word and two sentences, whether the word is used with the same sense in both sentence? Choose one from the option.\\Option: True, False} & \parbox{5cm}{word: touch\\He has a touch of rheumatism.\\He longed for the touch of her hand.} & False \\

\bottomrule
\end{tabular}

}
\caption{The example of the 15 downstream tasks for fine-tuning}
\label{tab:detailed_tasks2}
\end{table*}

\subsection{Task-wise Results On Continual Fine-tuning 15 Downstream Tasks}

We show the task-wise results of the continual fine-tuning experiments as in Table.\ref{tab:detail_full_first}\textasciitilde\ref{tab:detail_lora_last}, corresponding to each entry in Table.\ref{tab:continual_full} and Table.\ref{tab:continual_lora}. For instance, Table.\ref{tab:detail_full_first} elaborates the Baseline entry in Table.\ref{tab:continual_full} by including performance of the previous 14 tasks, rather than only listing the final result of the last task. 
The first column indicates the current learning task, while the remaining columns show evaluation metrics for previously learned tasks and the resulted MMLU and F1 Avg at each learning step.

\begin{table*}[htbp]
\centering
\setlength{\tabcolsep}{2pt}
\resizebox{\linewidth}{!}{

\begin{tabular}{lcccccccccccccccccc}
\toprule
Curr.Task &  F1 Avg &    MMLU & Tasks Avg &    yelp &  amazon &    MNLI &      CB &    COPA &     QQP &     RTE &    IMDB &   SST-2 & dbpedia &  agnews &   yahoo & MultiRC &  BoolQA &     WiC \\
\midrule
     yelp & 63.8440 & 62.0571 &   65.7368 & 65.7368 & -       & -       & -       & -       & -       & -       & -       & -       & -       & -       & -       & -       & -       & -       \\
   amazon & 60.9339 & 58.1699 &   63.9737 & 63.4605 & 64.4868 & -       & -       & -       & -       & -       & -       & -       & -       & -       & -       & -       & -       & -       \\
     MNLI & 58.7423 & 55.0740 &   62.9342 & 53.8158 & 49.8158 & 85.1711 & -       & -       & -       & -       & -       & -       & -       & -       & -       & -       & -       & -       \\
       CB & 57.8043 & 55.3492 &   60.4874 & 51.4737 & 47.4605 & 82.2500 & 98.2143 & -       & -       & -       & -       & -       & -       & -       & -       & -       & -       & -       \\
     COPA & 31.0482 & 52.8724 &   21.9768 & 36.4605 & 28.6711 &  0.0000 &  0.0000 & 95.0000 & -       & -       & -       & -       & -       & -       & -       & -       & -       & -       \\
      QQP & 57.5412 & 53.5604 &   62.1613 & 54.8289 & 55.7500 & 52.6447 & 51.7857 & 98.0000 & 85.0263 & -       & -       & -       & -       & -       & -       & -       & -       & -       \\
      RTE & 57.4316 & 51.8060 &   64.4277 & 55.0789 & 56.0921 & 61.4211 & 87.5000 & 97.0000 & 83.5921 & 89.8917 & -       & -       & -       & -       & -       & -       & -       & -       \\
\bottomrule
\end{tabular}

}
\caption{The task-wise performance (\%) of Baseline under full-parameter setting}
\label{tab:detail_full_first}
\end{table*}

\begin{table*}[htbp]
\centering

\setlength{\tabcolsep}{2pt}
\resizebox{\linewidth}{!}{

\begin{tabular}{lcccccccccccccccccc}
\toprule
Curr.Task &  F1 Avg &    MMLU & Tasks Avg &    yelp &  amazon &    MNLI &      CB &    COPA &     QQP &     RTE &    IMDB &   SST-2 & dbpedia &  agnews &   yahoo & MultiRC &  BoolQA &     WiC \\
\midrule
     yelp & 64.6890 & 62.9859 &   66.4868 & 66.4868 & -       & -       & -       & -       & -       & -       & -       & -       & -       & -       & -       & -       & -       & -       \\
   amazon & 62.8706 & 61.1971 &   64.6382 & 64.6579 & 64.6184 & -       & -       & -       & -       & -       & -       & -       & -       & -       & -       & -       & -       & -       \\
     MNLI & 59.6392 & 56.0028 &   63.7807 & 55.9211 & 50.0395 & 85.3816 & -       & -       & -       & -       & -       & -       & -       & -       & -       & -       & -       & -       \\
       CB & 57.6580 & 53.4228 &   62.6225 & 56.3026 & 50.5132 & 80.8158 & 94.6429 & -       & -       & -       & -       & -       & -       & -       & -       & -       & -       & -       \\
     COPA & 46.5305 & 57.0003 &   39.3100 & 51.5395 & 45.2763 & 20.5921 &  8.9286 & 96.0000 & -       & -       & -       & -       & -       & -       & -       & -       & -       & -       \\
      QQP & 57.5918 & 55.6244 &   59.7035 & 58.6579 & 54.3816 & 39.7763 & 48.2143 & 97.0000 & 85.5921 & -       & -       & -       & -       & -       & -       & -       & -       & -       \\
      RTE & 58.7280 & 54.3172 &   63.9185 & 56.2632 & 50.3553 & 62.8158 & 83.9286 & 95.0000 & 84.7368 & 89.8917 & -       & -       & -       & -       & -       & -       & -       & -       \\
\bottomrule
\end{tabular}

}
\caption{The task-wise performance (\%) of Baseline\textsuperscript{R} under full-parameter setting}
\end{table*}

\begin{table*}[htbp]
\centering

\setlength{\tabcolsep}{2pt}
\resizebox{\linewidth}{!}{

\begin{tabular}{lcccccccccccccccccc}
\toprule
Curr.Task &  F1 Avg &    MMLU & Tasks Avg &    yelp &  amazon &    MNLI &      CB &    COPA &     QQP &     RTE &    IMDB &   SST-2 & dbpedia &  agnews &   yahoo & MultiRC &  BoolQA &     WiC \\
\midrule
     yelp & 64.0879 & 62.4011 &   65.8684 & 65.8684 & -       & -       & -       & -       & -       & -       & -       & -       & -       & -       & -       & -       & -       & -       \\
   amazon & 62.5326 & 60.6467 &   64.5395 & 64.8158 & 64.2632 & -       & -       & -       & -       & -       & -       & -       & -       & -       & -       & -       & -       & -       \\
     MNLI & 57.3472 & 58.2387 &   56.4825 & 46.3421 & 37.6447 & 85.4605 & -       & -       & -       & -       & -       & -       & -       & -       & -       & -       & -       & -       \\
       CB & 53.2286 & 58.1355 &   49.0856 & 39.6974 & 31.0263 & 76.2368 & 89.2857 & -       & -       & -       & -       & -       & -       & -       & -       & -       & -       & -       \\
     COPA & 44.9626 & 57.7227 &   36.8226 & 58.0789 & 51.6974 &  0.2105 &  0.0000 & 94.0000 & -       & -       & -       & -       & -       & -       & -       & -       & -       & -       \\
      QQP & 61.5757 & 57.9291 &   65.7121 & 63.8158 & 61.5921 & 51.4342 & 80.3571 & 88.0000 & 85.6053 & -       & -       & -       & -       & -       & -       & -       & -       & -       \\
      RTE & 60.0372 & 54.9364 &   66.1823 & 61.6579 & 57.7763 & 60.7368 & 80.3571 & 85.0000 & 83.2895 & 91.3357 & -       & -       & -       & -       & -       & -       & -       & -       \\
\bottomrule
\end{tabular}

}
\caption{The task-wise performance (\%) of Baseline\textsuperscript{R} with BI under full-parameter setting}
\end{table*}

\begin{table*}[htbp]
\centering

\setlength{\tabcolsep}{2pt}
\resizebox{\linewidth}{!}{

\begin{tabular}{lcccccccccccccccccc}
\toprule
Curr.Task &  F1 Avg &    MMLU & Tasks Avg &    yelp &  amazon &    MNLI &       CB &    COPA &     QQP &     RTE &    IMDB &   SST-2 & dbpedia &  agnews &   yahoo & MultiRC &  BoolQA &     WiC \\
\midrule
     yelp & 64.8132 & 63.5707 &   66.1053 & 66.1053 & -       & -       &  -       & -       & -       & -       & -       & -       & -       & -       & -       & -       & -       & -       \\
   amazon & 62.3487 & 60.4403 &   64.3816 & 64.8158 & 63.9474 & -       &  -       & -       & -       & -       & -       & -       & -       & -       & -       & -       & -       & -       \\
     MNLI & 61.7917 & 56.2092 &   68.6053 & 63.0395 & 56.9211 & 85.8553 &  -       & -       & -       & -       & -       & -       & -       & -       & -       & -       & -       & -       \\
       CB & 61.9254 & 56.4499 &   68.5772 & 63.0526 & 57.6711 & 84.7763 & 100.0000 & -       & -       & -       & -       & -       & -       & -       & -       & -       & -       & -       \\
     COPA & 45.0773 & 56.1060 &   37.6721 & 55.6184 & 53.0789 &  3.8158 &   1.7857 & 96.0000 & -       & -       & -       & -       & -       & -       & -       & -       & -       & -       \\
      QQP & 57.0507 & 54.2484 &   60.1584 & 58.3289 & 57.7500 & 38.8158 &  57.1429 & 86.0000 & 85.4211 & -       & -       & -       & -       & -       & -       & -       & -       & -       \\
      RTE & 58.2697 & 52.9412 &   64.7910 & 57.5132 & 55.1184 & 61.5789 &  85.7143 & 87.0000 & 83.5658 & 90.6137 & -       & -       & -       & -       & -       & -       & -       & -       \\
\bottomrule
\end{tabular}

}
\caption{The task-wise performance (\%) of Baseline\textsuperscript{R}+KL under full-parameter setting}
\end{table*}

\begin{table*}[htbp]
\centering

\setlength{\tabcolsep}{2pt}
\resizebox{\linewidth}{!}{

\begin{tabular}{lcccccccccccccccccc}
\toprule
Curr.Task &  F1 Avg &    MMLU & Tasks Avg &    yelp &  amazon &    MNLI &      CB &    COPA &     QQP &     RTE &    IMDB &   SST-2 & dbpedia &  agnews &   yahoo & MultiRC &  BoolQA &     WiC \\
\midrule
     yelp & 64.3784 & 63.1235 &   65.6842 & 65.6842 & -       & -       & -       & -       & -       & -       & -       & -       & -       & -       & -       & -       & -       & -       \\
   amazon & 63.1231 & 62.0227 &   64.2632 & 64.5921 & 63.9342 & -       & -       & -       & -       & -       & -       & -       & -       & -       & -       & -       & -       & -       \\
     MNLI & 62.1276 & 59.8555 &   64.5789 & 55.9868 & 52.7237 & 85.0263 & -       & -       & -       & -       & -       & -       & -       & -       & -       & -       & -       & -       \\
       CB & 58.7442 & 59.9587 &   57.5779 & 50.6974 & 45.8816 & 75.8816 & 94.6429 & -       & -       & -       & -       & -       & -       & -       & -       & -       & -       & -       \\
     COPA & 42.4836 & 59.5803 &   33.0110 & 50.8816 & 47.4342 &  0.1053 &  0.0000 & 98.0000 & -       & -       & -       & -       & -       & -       & -       & -       & -       & -       \\
      QQP & 58.3575 & 53.6636 &   63.9514 & 60.4342 & 53.7500 & 55.7105 & 80.3571 & 76.0000 & 85.6316 & -       & -       & -       & -       & -       & -       & -       & -       & -       \\
      RTE & 58.1002 & 52.1156 &   65.6375 & 60.3421 & 53.6184 & 62.9605 & 87.5000 & 85.0000 & 84.3158 & 90.2527 & -       & -       & -       & -       & -       & -       & -       & -       \\
\bottomrule
\end{tabular}

}
\caption{The task-wise performance (\%) of Baseline\textsuperscript{R}+KL with BI under full-parameter setting}
\end{table*}

\begin{table*}[htbp]
\centering

\setlength{\tabcolsep}{2pt}
\resizebox{\linewidth}{!}{

\begin{tabular}{lcccccccccccccccccc}
\toprule
Curr.Task &  F1 Avg &    MMLU & Tasks Avg &    yelp &  amazon &    MNLI &      CB &    COPA &     QQP &     RTE &    IMDB &   SST-2 & dbpedia &  agnews &   yahoo & MultiRC &  BoolQA &     WiC \\
\midrule
     yelp & 65.3544 & 64.9123 &   65.8026 & 65.8026 & -       & -       & -       & -       & -       & -       & -       & -       & -       & -       & -       & -       & -       & -       \\
   amazon & 63.3547 & 62.9515 &   63.7632 & 63.8421 & 63.6842 & -       & -       & -       & -       & -       & -       & -       & -       & -       & -       & -       & -       & -       \\
     MNLI & 65.8955 & 61.4035 &   71.0965 & 64.9605 & 62.8289 & 85.5000 & -       & -       & -       & -       & -       & -       & -       & -       & -       & -       & -       & -       \\
       CB & 66.4280 & 62.1603 &   71.3248 & 65.1053 & 63.2105 & 85.4868 & 94.6429 & -       & -       & -       & -       & -       & -       & -       & -       & -       & -       & -       \\
     COPA & 55.2835 & 59.2363 &   51.8252 & 62.6447 & 61.8553 & 30.5395 & 33.9286 & 95.0000 & -       & -       & -       & -       & -       & -       & -       & -       & -       & -       \\
      QQP & 63.2466 & 57.3443 &   70.5033 & 63.4211 & 62.1579 & 69.8421 & 89.2857 & 91.0000 & 86.1842 & -       & -       & -       & -       & -       & -       & -       & -       & -       \\
      RTE & 62.3130 & 57.0691 &   68.6180 & 64.1974 & 61.3421 & 62.5789 & 87.5000 & 93.0000 & 85.1053 & 90.2527 & -       & -       & -       & -       & -       & -       & -       & -       \\
\bottomrule
\end{tabular}

}
\caption{The task-wise performance (\%) of Baseline\textsuperscript{R}+L1 under full-parameter setting}
\end{table*}

\begin{table*}[htbp]
\centering

\setlength{\tabcolsep}{2pt}
\resizebox{\linewidth}{!}{

\begin{tabular}{lcccccccccccccccccc}
\toprule
Curr.Task &  F1 Avg &    MMLU & Tasks Avg &    yelp &  amazon &    MNLI &      CB &    COPA &     QQP &     RTE &    IMDB &   SST-2 & dbpedia &  agnews &   yahoo & MultiRC &  BoolQA &     WiC \\
\midrule
     yelp & 65.1065 & 64.2243 &   66.0132 & 66.0132 & -       & -       & -       & -       & -       & -       & -       & -       & -       & -       & -       & -       & -       & -       \\
   amazon & 63.3069 & 62.7107 &   63.9145 & 64.2237 & 63.6053 & -       & -       & -       & -       & -       & -       & -       & -       & -       & -       & -       & -       & -       \\
     MNLI & 63.4374 & 57.7227 &   70.4079 & 64.7237 & 61.2237 & 85.2763 & -       & -       & -       & -       & -       & -       & -       & -       & -       & -       & -       & -       \\
       CB & 62.6949 & 56.6563 &   70.1741 & 64.6842 & 62.1711 & 83.5263 & 89.2857 & -       & -       & -       & -       & -       & -       & -       & -       & -       & -       & -       \\
     COPA & 44.5302 & 56.2779 &   36.8400 & 56.9605 & 53.0263 &  0.0000 &  0.0000 & 98.0000 & -       & -       & -       & -       & -       & -       & -       & -       & -       & -       \\
      QQP & 58.2426 & 52.1844 &   65.8921 & 60.2237 & 59.4737 & 57.6053 & 83.9286 & 92.0000 & 85.7895 & -       & -       & -       & -       & -       & -       & -       & -       & -       \\
      RTE & 56.6823 & 49.0196 &   67.1845 & 61.1711 & 59.5395 & 62.6974 & 80.3571 & 88.0000 & 84.0658 & 91.6968 & -       & -       & -       & -       & -       & -       & -       & -       \\
\bottomrule
\end{tabular}

}
\caption{The task-wise performance (\%) of Baseline\textsuperscript{R}+L1 with BI under full-parameter setting}
\end{table*}

\begin{table*}[htbp]
\centering

\setlength{\tabcolsep}{2pt}
\resizebox{\linewidth}{!}{

\begin{tabular}{lcccccccccccccccccc}
\toprule
Curr.Task &  F1 Avg &    MMLU & Tasks Avg &    yelp &  amazon &    MNLI &      CB &    COPA &     QQP &     RTE &    IMDB &   SST-2 & dbpedia &  agnews &   yahoo & MultiRC &  BoolQA &     WiC \\
\midrule
     yelp & 64.9877 & 64.6371 &   65.3421 & 65.3421 & -       & -       & -       & -       & -       & -       & -       & -       & -       & -       & -       & -       & -       & -       \\
   amazon & 63.7353 & 63.8459 &   63.6250 & 63.9342 & 63.3158 & -       & -       & -       & -       & -       & -       & -       & -       & -       & -       & -       & -       & -       \\
     MNLI & 65.0210 & 59.6491 &   71.4561 & 65.0263 & 63.1579 & 86.1842 & -       & -       & -       & -       & -       & -       & -       & -       & -       & -       & -       & -       \\
       CB & 65.5054 & 60.8187 &   70.9748 & 65.1184 & 62.9474 & 84.6842 & 94.6429 & -       & -       & -       & -       & -       & -       & -       & -       & -       & -       & -       \\
     COPA & 55.3894 & 59.1675 &   52.0648 & 62.9605 & 61.6316 & 31.1579 & 30.3571 & 98.0000 & -       & -       & -       & -       & -       & -       & -       & -       & -       & -       \\
      QQP & 63.0276 & 57.6883 &   69.4561 & 63.4868 & 61.2368 & 66.7237 & 85.7143 & 93.0000 & 85.9474 & -       & -       & -       & -       & -       & -       & -       & -       & -       \\
      RTE & 61.9077 & 57.2067 &   67.4505 & 63.8816 & 59.6184 & 61.7237 & 87.5000 & 95.0000 & 83.1316 & 93.1408 & -       & -       & -       & -       & -       & -       & -       & -       \\
\bottomrule
\end{tabular}

}
\caption{The task-wise performance (\%) of Baseline\textsuperscript{R}+L2 under full-parameter setting}
\end{table*}

\begin{table*}[htbp]
\centering

\setlength{\tabcolsep}{2pt}
\resizebox{\linewidth}{!}{

\begin{tabular}{lcccccccccccccccccc}
\toprule
Curr.Task &  F1 Avg &    MMLU & Tasks Avg &    yelp &  amazon &    MNLI &      CB &    COPA &     QQP &     RTE &    IMDB &   SST-2 & dbpedia &  agnews &   yahoo & MultiRC &  BoolQA &     WiC \\
\midrule
     yelp & 64.9750 & 64.1555 &   65.8158 & 65.8158 & -       & -       & -       & -       & -       & -       & -       & -       & -       & -       & -       & -       & -       & -       \\
   amazon & 63.8210 & 63.7083 &   63.9342 & 64.0263 & 63.8421 & -       & -       & -       & -       & -       & -       & -       & -       & -       & -       & -       & -       & -       \\
     MNLI & 65.9386 & 62.2291 &   70.1184 & 64.8684 & 60.8684 & 84.6184 & -       & -       & -       & -       & -       & -       & -       & -       & -       & -       & -       & -       \\
       CB & 65.0262 & 60.8187 &   69.8591 & 65.7105 & 61.1579 & 82.5132 & 96.4286 & -       & -       & -       & -       & -       & -       & -       & -       & -       & -       & -       \\
     COPA & 50.7358 & 59.8555 &   44.0277 & 61.6711 & 60.5395 &  9.4737 &  3.5714 & 97.0000 & -       & -       & -       & -       & -       & -       & -       & -       & -       & -       \\
      QQP & 62.3038 & 56.6219 &   69.2532 & 63.4868 & 61.2895 & 65.8289 & 83.9286 & 91.0000 & 86.0132 & -       & -       & -       & -       & -       & -       & -       & -       & -       \\
      RTE & 61.5376 & 56.1404 &   68.0829 & 63.6974 & 60.0921 & 62.8947 & 87.5000 & 93.0000 & 84.3816 & 89.8917 & -       & -       & -       & -       & -       & -       & -       & -       \\
\bottomrule
\end{tabular}

}
\caption{The task-wise performance (\%) of Baseline\textsuperscript{R}+L2 with BI under full-parameter setting}
\end{table*}

\begin{table*}[htbp]
\centering

\setlength{\tabcolsep}{2pt}
\resizebox{\linewidth}{!}{

\begin{tabular}{lcccccccccccccccccc}
\toprule
Curr.Task &  F1 Avg &    MMLU & Tasks Avg &    yelp &  amazon &    MNLI &      CB &    COPA &     QQP &     RTE &    IMDB &   SST-2 & dbpedia &  agnews &   yahoo & MultiRC &  BoolQA &     WiC \\
\midrule
     yelp & 65.6759 & 66.0131 &   65.3421 & 65.3421 & -       & -       & -       & -       & -       & -       & -       & -       & -       & -       & -       & -       & -       & -       \\
   amazon & 64.7975 & 65.0843 &   64.5132 & 64.8421 & 64.1842 & -       & -       & -       & -       & -       & -       & -       & -       & -       & -       & -       & -       & -       \\
     MNLI & 66.9683 & 64.7059 &   69.3947 & 62.9474 & 60.8684 & 84.3684 & -       & -       & -       & -       & -       & -       & -       & -       & -       & -       & -       & -       \\
       CB & 66.9523 & 64.8091 &   69.2422 & 62.8553 & 60.9474 & 83.7763 & 89.2857 & -       & -       & -       & -       & -       & -       & -       & -       & -       & -       & -       \\
     COPA & 65.2217 & 63.6395 &   66.8845 & 63.0132 & 61.5000 & 75.6184 & 87.5000 & 95.0000 & -       & -       & -       & -       & -       & -       & -       & -       & -       & -       \\
      QQP & 67.0591 & 64.4307 &   69.9110 & 62.7500 & 60.3289 & 71.1711 & 87.5000 & 97.0000 & 84.9079 & -       & -       & -       & -       & -       & -       & -       & -       & -       \\
      RTE & 64.1356 & 62.2979 &   66.0850 & 61.8684 & 59.4605 & 61.4605 & 85.7143 & 96.0000 & 80.1842 & 88.8087 & -       & -       & -       & -       & -       & -       & -       & -       \\
\bottomrule
\end{tabular}

}
\caption{The task-wise performance (\%) of Baseline\textsuperscript{R}+L1 (w=100) under full-parameter setting}
\end{table*}

\begin{table*}[htbp]
\centering

\setlength{\tabcolsep}{2pt}
\resizebox{\linewidth}{!}{

\begin{tabular}{lcccccccccccccccccc}
\toprule
Curr.Task &  F1 Avg &    MMLU & Tasks Avg &    yelp &  amazon &    MNLI &      CB &    COPA &     QQP &     RTE &    IMDB &   SST-2 & dbpedia &  agnews &   yahoo & MultiRC &  BoolQA &     WiC \\
\midrule
     yelp & 66.3682 & 66.4603 &   66.2763 & 66.2763 & -       & -       & -       & -       & -       & -       & -       & -       & -       & -       & -       & -       & -       & -       \\
   amazon & 64.9774 & 65.1531 &   64.8026 & 65.0000 & 64.6053 & -       & -       & -       & -       & -       & -       & -       & -       & -       & -       & -       & -       & -       \\
     MNLI & 65.9586 & 64.7403 &   67.2237 & 59.8026 & 58.0263 & 83.8421 & -       & -       & -       & -       & -       & -       & -       & -       & -       & -       & -       & -       \\
       CB & 65.5301 & 64.3963 &   66.7046 & 60.2500 & 58.0395 & 81.6316 & 92.8571 & -       & -       & -       & -       & -       & -       & -       & -       & -       & -       & -       \\
     COPA & 62.8714 & 62.8139 &   62.9291 & 59.5526 & 57.2895 & 71.3158 & 87.5000 & 97.0000 & -       & -       & -       & -       & -       & -       & -       & -       & -       & -       \\
      QQP & 64.7986 & 63.4675 &   66.1867 & 55.1974 & 52.2368 & 71.9342 & 92.8571 & 95.0000 & 84.8026 & -       & -       & -       & -       & -       & -       & -       & -       & -       \\
      RTE & 63.7388 & 62.8483 &   64.6548 & 56.9474 & 55.0132 & 61.5789 & 91.0714 & 95.0000 & 83.6974 & 86.2816 & -       & -       & -       & -       & -       & -       & -       & -       \\
\bottomrule
\end{tabular}

}
\caption{The task-wise performance (\%) of Baseline\textsuperscript{R}+L1 (w=100) with BI under full-parameter setting}
\end{table*}

\begin{table*}[htbp]
\centering

\setlength{\tabcolsep}{2pt}
\resizebox{\linewidth}{!}{

\begin{tabular}{lcccccccccccccccccc}
\toprule
Curr.Task &  F1 Avg &    MMLU & Tasks Avg &    yelp &  amazon &    MNLI &      CB &    COPA &     QQP &     RTE &    IMDB &   SST-2 & dbpedia &  agnews &   yahoo & MultiRC &  BoolQA &     WiC \\
\midrule
     yelp & 65.7533 & 65.8755 &   65.6316 & 65.6316 & -       & -       & -       & -       & -       & -       & -       & -       & -       & -       & -       & -       & -       & -       \\
   amazon & 64.7048 & 65.1187 &   64.2961 & 64.1053 & 64.4868 & -       & -       & -       & -       & -       & -       & -       & -       & -       & -       & -       & -       & -       \\
     MNLI & 66.8373 & 64.1211 &   69.7939 & 63.5132 & 60.4474 & 85.4211 & -       & -       & -       & -       & -       & -       & -       & -       & -       & -       & -       & -       \\
       CB & 66.9012 & 64.2243 &   69.8110 & 63.3158 & 60.6579 & 85.3158 & 89.2857 & -       & -       & -       & -       & -       & -       & -       & -       & -       & -       & -       \\
     COPA & 60.6423 & 63.2955 &   58.2026 & 62.5395 & 60.6447 & 51.1316 & 28.5714 & 97.0000 & -       & -       & -       & -       & -       & -       & -       & -       & -       & -       \\
      QQP & 66.4575 & 61.6099 &   72.1331 & 62.4211 & 59.5263 & 80.2105 & 89.2857 & 94.0000 & 85.9605 & -       & -       & -       & -       & -       & -       & -       & -       & -       \\
      RTE & 62.4516 & 59.9587 &   65.1607 & 60.6447 & 56.4605 & 62.1974 & 80.3571 & 90.0000 & 80.0000 & 89.8917 & -       & -       & -       & -       & -       & -       & -       & -       \\
\bottomrule
\end{tabular}

}
\caption{The task-wise performance (\%) of Baseline\textsuperscript{R}+L2 (w=100) under full-parameter setting}
\end{table*}

\begin{table*}[htbp]
\centering

\setlength{\tabcolsep}{2pt}
\resizebox{\linewidth}{!}{

\begin{tabular}{lcccccccccccccccccc}
\toprule
Curr.Task &  F1 Avg &    MMLU & Tasks Avg &    yelp &  amazon &    MNLI &      CB &    COPA &     QQP &     RTE &    IMDB &   SST-2 & dbpedia &  agnews &   yahoo & MultiRC &  BoolQA &     WiC \\
\midrule
     yelp & 65.7186 & 65.9787 &   65.4605 & 65.4605 & -       & -       & -       & -       & -       & -       & -       & -       & -       & -       & -       & -       & -       & -       \\
   amazon & 64.9316 & 65.6691 &   64.2105 & 64.6053 & 63.8158 & -       & -       & -       & -       & -       & -       & -       & -       & -       & -       & -       & -       & -       \\
     MNLI & 63.5951 & 64.6027 &   62.6184 & 53.8684 & 48.1842 & 85.8026 & -       & -       & -       & -       & -       & -       & -       & -       & -       & -       & -       & -       \\
       CB & 63.7175 & 64.6371 &   62.8238 & 56.4079 & 50.9079 & 80.9211 & 94.6429 & -       & -       & -       & -       & -       & -       & -       & -       & -       & -       & -       \\
     COPA & 32.8576 & 61.4379 &   22.4255 & 36.8947 & 29.5658 &  0.0000 &  0.0000 & 97.0000 & -       & -       & -       & -       & -       & -       & -       & -       & -       & -       \\
      QQP & 63.9195 & 61.2315 &   66.8543 & 64.3421 & 61.0526 & 56.2500 & 51.7857 & 90.0000 & 85.5789 & -       & -       & -       & -       & -       & -       & -       & -       & -       \\
      RTE & 61.8861 & 57.6539 &   66.7888 & 63.3684 & 57.4605 & 62.4868 & 83.9286 & 95.0000 & 82.5658 & 88.0866 & -       & -       & -       & -       & -       & -       & -       & -       \\
\bottomrule
\end{tabular}

}
\caption{The task-wise performance (\%) of Baseline\textsuperscript{R}+L2 (w=100) with BI under full-parameter setting}
\end{table*}

\begin{table*}[htbp]
\centering

\setlength{\tabcolsep}{2pt}
\resizebox{\linewidth}{!}{

\begin{tabular}{lcccccccccccccccccc}
\toprule
Curr.Task &  F1 Avg &    MMLU & Tasks Avg &    yelp &  amazon &    MNLI &      CB &    COPA &     QQP &     RTE &    IMDB &   SST-2 & dbpedia &  agnews &   yahoo & MultiRC &  BoolQA &     WiC \\
\midrule
     yelp & 65.8252 & 65.7035 &   65.9474 & 65.9474 & -       & -       & -       & -       & -       & -       & -       & -       & -       & -       & -       & -       & -       & -       \\
   amazon & 64.7036 & 65.0155 &   64.3947 & 64.9868 & 63.8026 & -       & -       & -       & -       & -       & -       & -       & -       & -       & -       & -       & -       & -       \\
     MNLI & 66.9695 & 62.7795 &   71.7588 & 65.2105 & 64.1579 & 85.9079 & -       & -       & -       & -       & -       & -       & -       & -       & -       & -       & -       & -       \\
       CB & 66.3947 & 62.0227 &   71.4298 & 65.1842 & 64.4211 & 84.5132 & 94.6429 & -       & -       & -       & -       & -       & -       & -       & -       & -       & -       & -       \\
     COPA & 51.3578 & 59.8555 &   44.9730 & 63.7500 & 63.3158 &  7.4868 &  0.0000 & 98.0000 & -       & -       & -       & -       & -       & -       & -       & -       & -       & -       \\
      QQP & 66.0079 & 60.8531 &   72.1168 & 64.2895 & 63.2763 & 74.3158 & 85.7143 & 98.0000 & 86.1447 & -       & -       & -       & -       & -       & -       & -       & -       & -       \\
      RTE & 64.3613 & 60.4059 &   68.8710 & 63.8816 & 62.2895 & 62.9605 & 87.5000 & 98.0000 & 85.0000 & 91.6968 & -       & -       & -       & -       & -       & -       & -       & -       \\
\bottomrule
\end{tabular}

}
\caption{The task-wise performance (\%) of Baseline\textsuperscript{R}+L1 (w=dy) under full-parameter setting}
\end{table*}

\begin{table*}[htbp]
\centering

\setlength{\tabcolsep}{2pt}
\resizebox{\linewidth}{!}{

\begin{tabular}{lcccccccccccccccccc}
\toprule
Curr.Task &  F1 Avg &    MMLU & Tasks Avg &    yelp &  amazon &    MNLI &      CB &    COPA &     QQP &     RTE &    IMDB &   SST-2 & dbpedia &  agnews &   yahoo & MultiRC &  BoolQA &     WiC \\
\midrule
     yelp & 65.7648 & 65.3251 &   66.2105 & 66.2105 & -       & -       & -       & -       & -       & -       & -       & -       & -       & -       & -       & -       & -       & -       \\
   amazon & 63.7685 & 63.7083 &   63.8289 & 64.4079 & 63.2500 & -       & -       & -       & -       & -       & -       & -       & -       & -       & -       & -       & -       & -       \\
     MNLI & 64.3991 & 61.2315 &   67.9123 & 61.3421 & 57.0789 & 85.3158 & -       & -       & -       & -       & -       & -       & -       & -       & -       & -       & -       & -       \\
       CB & 63.8369 & 60.6811 &   67.3390 & 61.7763 & 57.7500 & 82.2763 & 96.4286 & -       & -       & -       & -       & -       & -       & -       & -       & -       & -       & -       \\
     COPA &  9.8956 & 58.8923 &    5.4016 & 11.6974 &  3.3421 &  0.0000 &  0.0000 & 97.0000 & -       & -       & -       & -       & -       & -       & -       & -       & -       & -       \\
      QQP & 62.0373 & 56.2092 &   69.2139 & 64.6579 & 61.4868 & 64.6447 & 87.5000 & 92.0000 & 85.6316 & -       & -       & -       & -       & -       & -       & -       & -       & -       \\
      RTE & 58.2216 & 53.8700 &   63.3380 & 63.5789 & 58.4737 & 62.1974 & 87.5000 & 92.0000 & 67.5658 & 90.2527 & -       & -       & -       & -       & -       & -       & -       & -       \\
\bottomrule
\end{tabular}

}
\caption{The task-wise performance (\%) of Baseline\textsuperscript{R}+L1 (w=dy) with BI under full-parameter setting}
\end{table*}

\begin{table*}[htbp]
\centering

\setlength{\tabcolsep}{2pt}
\resizebox{\linewidth}{!}{

\begin{tabular}{lcccccccccccccccccc}
\toprule
Curr.Task &  F1 Avg &    MMLU & Tasks Avg &    yelp &  amazon &    MNLI &      CB &    COPA &     QQP &     RTE &    IMDB &   SST-2 & dbpedia &  agnews &   yahoo & MultiRC &  BoolQA &     WiC \\
\midrule
     yelp & 66.3052 & 66.1507 &   66.4605 & 66.4605 & -       & -       & -       & -       & -       & -       & -       & -       & -       & -       & -       & -       & -       & -       \\
   amazon & 64.8726 & 64.6027 &   65.1447 & 65.4737 & 64.8158 & -       & -       & -       & -       & -       & -       & -       & -       & -       & -       & -       & -       & -       \\
     MNLI & 66.7933 & 63.1579 &   70.8728 & 64.1711 & 63.6316 & 84.8158 & -       & -       & -       & -       & -       & -       & -       & -       & -       & -       & -       & -       \\
       CB & 66.7035 & 62.9515 &   70.9310 & 64.4342 & 63.4605 & 84.7237 & 94.6429 & -       & -       & -       & -       & -       & -       & -       & -       & -       & -       & -       \\
     COPA & 49.6547 & 59.8555 &   42.4246 & 63.1974 & 62.5526 &  1.0921 &  3.5714 & 97.0000 & -       & -       & -       & -       & -       & -       & -       & -       & -       & -       \\
      QQP & 63.6291 & 57.3099 &   71.5146 & 62.9737 & 61.9868 & 74.7763 & 89.2857 & 96.0000 & 85.8684 & -       & -       & -       & -       & -       & -       & -       & -       & -       \\
      RTE & 61.9088 & 57.0347 &   67.6937 & 63.0000 & 60.3684 & 62.1711 & 83.9286 & 97.0000 & 83.9868 & 88.0866 & -       & -       & -       & -       & -       & -       & -       & -       \\
\bottomrule
\end{tabular}

}
\caption{The task-wise performance (\%) of Baseline\textsuperscript{R}+L2 (w=dy) under full-parameter setting}
\end{table*}

\begin{table*}[htbp]
\centering

\setlength{\tabcolsep}{2pt}
\resizebox{\linewidth}{!}{

\begin{tabular}{lcccccccccccccccccc}
\toprule
Curr.Task &  F1 Avg &    MMLU & Tasks Avg &    yelp &  amazon &    MNLI &      CB &    COPA &     QQP &     RTE &    IMDB &   SST-2 & dbpedia &  agnews &   yahoo & MultiRC &  BoolQA &     WiC \\
\midrule
     yelp & 65.7911 & 65.4283 &   66.1579 & 66.1579 & -       & -       & -       & -       & -       & -       & -       & -       & -       & -       & -       & -       & -       & -       \\
   amazon & 63.3645 & 62.9515 &   63.7829 & 64.0658 & 63.5000 & -       & -       & -       & -       & -       & -       & -       & -       & -       & -       & -       & -       & -       \\
     MNLI & 67.1038 & 63.7083 &   70.8816 & 65.2763 & 62.0000 & 85.3684 & -       & -       & -       & -       & -       & -       & -       & -       & -       & -       & -       & -       \\
       CB & 65.9670 & 62.9515 &   69.2860 & 64.7237 & 61.6579 & 81.2632 & 98.2143 & -       & -       & -       & -       & -       & -       & -       & -       & -       & -       & -       \\
     COPA & 46.6359 & 58.6859 &   38.6914 & 60.0789 & 55.5132 &  0.0000 &  0.0000 & 97.0000 & -       & -       & -       & -       & -       & -       & -       & -       & -       & -       \\
      QQP & 61.3321 & 55.8996 &   67.9343 & 64.5395 & 63.0263 & 57.8289 & 62.5000 & 92.0000 & 86.0658 & -       & -       & -       & -       & -       & -       & -       & -       & -       \\
      RTE & 59.3602 & 52.3564 &   68.5272 & 64.0000 & 61.6316 & 62.6447 & 83.9286 & 89.0000 & 84.7500 & 87.7256 & -       & -       & -       & -       & -       & -       & -       & -       \\
\bottomrule
\end{tabular}

}
\caption{The task-wise performance (\%) of Baseline\textsuperscript{R}+L2 (w=dy) with BI under full-parameter setting}
\end{table*}

\begin{table*}[htbp]
\centering

\setlength{\tabcolsep}{2pt}
\resizebox{\linewidth}{!}{

\begin{tabular}{lcccccccccccccccccc}
\toprule
Curr.Task &  F1 Avg &    MMLU & Tasks Avg &    yelp &  amazon &    MNLI &      CB &    COPA &     QQP &     RTE &    IMDB &   SST-2 & dbpedia &  agnews &   yahoo & MultiRC &  BoolQA &     WiC \\
\midrule
     yelp & 65.4521 & 65.0155 &   65.8947 & 65.8947 & -       & -       & -       & -       & -       & -       & -       & -       & -       & -       & -       & -       & -       & -       \\
   amazon & 63.8612 & 63.8803 &   63.8421 & 63.9868 & 63.6974 & -       & -       & -       & -       & -       & -       & -       & -       & -       & -       & -       & -       & -       \\
     MNLI & 65.4479 & 61.3003 &   70.1974 & 62.5658 & 61.6316 & 86.3947 & -       & -       & -       & -       & -       & -       & -       & -       & -       & -       & -       & -       \\
       CB & 65.3685 & 60.9219 &   70.5154 & 63.5000 & 61.9079 & 86.0000 & 89.2857 & -       & -       & -       & -       & -       & -       & -       & -       & -       & -       & -       \\
     COPA & 52.6889 & 58.0667 &   48.2227 & 60.8947 & 60.3816 & 23.0263 & 10.7143 & 97.0000 & -       & -       & -       & -       & -       & -       & -       & -       & -       & -       \\
      QQP & 63.0186 & 59.6147 &   66.8347 & 61.3553 & 59.4605 & 59.8421 & 80.3571 & 92.0000 & 86.2500 & -       & -       & -       & -       & -       & -       & -       & -       & -       \\
      RTE & 62.5993 & 58.5483 &   67.2526 & 62.0526 & 58.6842 & 62.3553 & 82.1429 & 96.0000 & 84.5658 & 90.9747 & -       & -       & -       & -       & -       & -       & -       & -       \\
\bottomrule
\end{tabular}

}
\caption{The task-wise performance (\%) of Baseline\textsuperscript{R}+TM under full-parameter setting}
\end{table*}

\begin{table*}[htbp]
\centering

\setlength{\tabcolsep}{2pt}
\resizebox{\linewidth}{!}{

\begin{tabular}{lcccccccccccccccccc}
\toprule
Curr.Task &  F1 Avg &    MMLU & Tasks Avg &    yelp &  amazon &    MNLI &      CB &    COPA &     QQP &     RTE &    IMDB &   SST-2 & dbpedia &  agnews &   yahoo & MultiRC &  BoolQA &     WiC \\
\midrule
     yelp & 65.1574 & 64.4995 &   65.8289 & 65.8289 & -       & -       & -       & -       & -       & -       & -       & -       & -       & -       & -       & -       & -       & -       \\
   amazon & 63.1013 & 61.8163 &   64.4408 & 64.9605 & 63.9211 & -       & -       & -       & -       & -       & -       & -       & -       & -       & -       & -       & -       & -       \\
     MNLI & 62.4140 & 57.4819 &   68.2719 & 61.7895 & 57.4737 & 85.5526 & -       & -       & -       & -       & -       & -       & -       & -       & -       & -       & -       & -       \\
       CB & 63.0028 & 57.9635 &   69.0016 & 63.3289 & 59.3158 & 84.2237 & 87.5000 & -       & -       & -       & -       & -       & -       & -       & -       & -       & -       & -       \\
     COPA & 45.2508 & 58.9611 &   36.7137 & 56.6974 & 52.4605 &  0.4474 &  0.0000 & 98.0000 & -       & -       & -       & -       & -       & -       & -       & -       & -       & -       \\
      QQP & 61.5596 & 57.6539 &   66.0329 & 62.0526 & 60.5132 & 55.2895 & 83.9286 & 94.0000 & 85.7763 & -       & -       & -       & -       & -       & -       & -       & -       & -       \\
      RTE & 61.1728 & 55.7276 &   67.7975 & 63.3289 & 60.0395 & 62.6447 & 78.5714 & 95.0000 & 83.8289 & 92.7798 & -       & -       & -       & -       & -       & -       & -       & -       \\
\bottomrule
\end{tabular}

}
\caption{The task-wise performance (\%) of Baseline\textsuperscript{R}+TM with BI under full-parameter setting}
\end{table*}

\begin{table*}[htbp]
\centering

\setlength{\tabcolsep}{2pt}
\resizebox{\linewidth}{!}{

\begin{tabular}{lcccccccccccccccccc}
\toprule
Curr.Task &  F1 Avg &    MMLU & Tasks Avg &    yelp &  amazon &    MNLI &      CB &    COPA &     QQP &     RTE &    IMDB &   SST-2 & dbpedia &  agnews &   yahoo & MultiRC &  BoolQA &     WiC \\
\midrule
     yelp & 65.1828 & 66.1851 &   64.2105 & 64.2105 & -       & -       & -       & -       & -       & -       & -       & -       & -       & -       & -       & -       & -       & -       \\
   amazon & 64.2106 & 65.2219 &   63.2303 & 63.6316 & 62.8289 & -       & -       & -       & -       & -       & -       & -       & -       & -       & -       & -       & -       & -       \\
     MNLI & 67.0835 & 64.5683 &   69.8026 & 63.6316 & 61.7368 & 84.0395 & -       & -       & -       & -       & -       & -       & -       & -       & -       & -       & -       & -       \\
       CB & 67.1489 & 64.4995 &   70.0254 & 63.8553 & 62.0395 & 84.0395 & 89.2857 & -       & -       & -       & -       & -       & -       & -       & -       & -       & -       & -       \\
     COPA & 67.3024 & 65.1531 &   69.5984 & 63.4079 & 61.6447 & 83.2632 & 89.2857 & 95.0000 & -       & -       & -       & -       & -       & -       & -       & -       & -       & -       \\
      QQP & 67.5704 & 64.9467 &   70.4150 & 61.0658 & 59.5658 & 76.6974 & 87.5000 & 95.0000 & 83.8816 & -       & -       & -       & -       & -       & -       & -       & -       & -       \\
      RTE & 65.2070 & 64.9123 &   65.5045 & 60.9605 & 57.7368 & 61.0921 & 83.9286 & 94.0000 & 80.9868 & 85.5596 & -       & -       & -       & -       & -       & -       & -       & -       \\
\bottomrule
\end{tabular}

}
\caption{The task-wise performance (\%) of Baseline\textsuperscript{R}+TM (w=100) under full-parameter setting}
\end{table*}

\begin{table*}[htbp]
\centering

\setlength{\tabcolsep}{2pt}
\resizebox{\linewidth}{!}{

\begin{tabular}{lcccccccccccccccccc}
\toprule
Curr.Task &  F1 Avg &    MMLU & Tasks Avg &    yelp &  amazon &    MNLI &      CB &    COPA &     QQP &     RTE &    IMDB &   SST-2 & dbpedia &  agnews &   yahoo & MultiRC &  BoolQA &     WiC \\
\midrule
     yelp & 65.6749 & 66.3915 &   64.9737 & 64.9737 & -       & -       & -       & -       & -       & -       & -       & -       & -       & -       & -       & -       & -       & -       \\
   amazon & 64.9092 & 65.2563 &   64.5658 & 64.7895 & 64.3421 & -       & -       & -       & -       & -       & -       & -       & -       & -       & -       & -       & -       & -       \\
     MNLI & 65.8225 & 65.2219 &   66.4342 & 57.8684 & 57.0132 & 84.4211 & -       & -       & -       & -       & -       & -       & -       & -       & -       & -       & -       & -       \\
       CB & 64.7232 & 64.6715 &   64.7751 & 59.1711 & 57.8158 & 77.1184 & 94.6429 & -       & -       & -       & -       & -       & -       & -       & -       & -       & -       & -       \\
     COPA & 50.8783 & 63.3643 &   42.5030 & 57.7895 & 54.6053 & 14.6579 &  7.1429 & 97.0000 & -       & -       & -       & -       & -       & -       & -       & -       & -       & -       \\
      QQP & 65.1654 & 62.8827 &   67.6201 & 56.8158 & 54.8947 & 72.9868 & 91.0714 & 95.0000 & 85.2500 & -       & -       & -       & -       & -       & -       & -       & -       & -       \\
      RTE & 62.4651 & 62.4355 &   62.4947 & 56.0526 & 53.8158 & 61.1447 & 89.2857 & 89.0000 & 77.5263 & 87.0036 & -       & -       & -       & -       & -       & -       & -       & -       \\
\bottomrule
\end{tabular}

}
\caption{The task-wise performance (\%) of Baseline\textsuperscript{R}+TM (w=100) with BI  under full-parameter setting}
\end{table*}

\begin{table*}[htbp]
\centering

\setlength{\tabcolsep}{2pt}
\resizebox{\linewidth}{!}{

\begin{tabular}{lcccccccccccccccccc}
\toprule
Curr.Task &  F1 Avg &    MMLU & Tasks Avg &    yelp &  amazon &    MNLI &      CB &    COPA &     QQP &     RTE &    IMDB &   SST-2 & dbpedia &  agnews &   yahoo & MultiRC &  BoolQA &     WiC \\
\midrule
     yelp & 66.2360 & 66.0131 &   66.4605 & 66.4605 & -       & -       & -       & -       & -       & -       & -       & -       & -       & -       & -       & -       & -       & -       \\
   amazon & 65.2747 & 66.0131 &   64.5526 & 64.5526 & 64.5526 & -       & -       & -       & -       & -       & -       & -       & -       & -       & -       & -       & -       & -       \\
     MNLI & 67.1812 & 63.7771 &   70.9693 & 64.9737 & 63.7632 & 84.1711 & -       & -       & -       & -       & -       & -       & -       & -       & -       & -       & -       & -       \\
       CB & 67.4603 & 63.8459 &   71.5086 & 65.2105 & 64.0395 & 85.1316 & 91.0714 & -       & -       & -       & -       & -       & -       & -       & -       & -       & -       & -       \\
     COPA & 67.7171 & 64.1899 &   71.6545 & 65.3421 & 63.9342 & 85.2237 & 87.5000 & 98.0000 & -       & -       & -       & -       & -       & -       & -       & -       & -       & -       \\
      QQP & 68.4733 & 64.9811 &   72.3622 & 64.5526 & 62.3947 & 76.8289 & 87.5000 & 97.0000 & 85.2368 & -       & -       & -       & -       & -       & -       & -       & -       & -       \\
      RTE & 66.7533 & 64.9467 &   68.6634 & 64.5658 & 62.6053 & 62.5658 & 87.5000 & 97.0000 & 83.6579 & 89.1697 & -       & -       & -       & -       & -       & -       & -       & -       \\
\bottomrule
\end{tabular}

}
\caption{The task-wise performance (\%) of Baseline\textsuperscript{R}+TM (w=dy) under full-parameter setting}
\end{table*}

\begin{table*}[htbp]
\centering

\setlength{\tabcolsep}{2pt}
\resizebox{\linewidth}{!}{

\begin{tabular}{lcccccccccccccccccc}
\toprule
Curr.Task &  F1 Avg &    MMLU & Tasks Avg &    yelp &  amazon &    MNLI &      CB &    COPA &     QQP &     RTE &    IMDB &   SST-2 & dbpedia &  agnews &   yahoo & MultiRC &  BoolQA &     WiC \\
\midrule
     yelp & 66.0494 & 65.7723 &   66.3289 & 66.3289 & -       & -       & -       & -       & -       & -       & -       & -       & -       & -       & -       & -       & -       & -       \\
   amazon & 64.0687 & 63.4331 &   64.7171 & 64.7632 & 64.6711 & -       & -       & -       & -       & -       & -       & -       & -       & -       & -       & -       & -       & -       \\
     MNLI & 67.5616 & 64.6371 &   70.7632 & 65.3816 & 62.2500 & 84.6579 & -       & -       & -       & -       & -       & -       & -       & -       & -       & -       & -       & -       \\
       CB & 66.9329 & 64.0179 &   70.1260 & 64.9737 & 62.1842 & 83.0132 & 98.2143 & -       & -       & -       & -       & -       & -       & -       & -       & -       & -       & -       \\
     COPA & 49.1982 & 58.4795 &   42.4595 & 60.0000 & 59.6974 &  7.2237 &  8.9286 & 96.0000 & -       & -       & -       & -       & -       & -       & -       & -       & -       & -       \\
      QQP & 65.3409 & 61.7819 &   69.3350 & 64.3421 & 62.7500 & 63.8421 & 62.5000 & 97.0000 & 86.0921 & -       & -       & -       & -       & -       & -       & -       & -       & -       \\
      RTE & 63.4401 & 59.8211 &   67.5251 & 62.7763 & 59.5658 & 63.1053 & 89.2857 & 96.0000 & 83.3158 & 89.5307 & -       & -       & -       & -       & -       & -       & -       & -       \\
\bottomrule
\end{tabular}

}
\caption{The task-wise performance (\%) of Baseline\textsuperscript{R}+TM (w=dy) with BI  under full-parameter setting}
\end{table*}
\begin{table*}[ht]
\centering
\setlength{\tabcolsep}{2pt}
\resizebox{\linewidth}{!}{

\begin{tabular}{lcccccccccccccccccc}
\toprule
Curr.Task &  F1 Avg &    MMLU & Tasks Avg &    yelp &  amazon &    MNLI &      CB &    COPA &     QQP &     RTE &    IMDB &   SST-2 & dbpedia &  agnews &   yahoo & MultiRC &  BoolQA &     WiC \\
\midrule
     yelp & 63.5458 & 63.3299 &   63.7632 & 63.7632 & -       & -       & -       & -       & -       & -       & -       & -       & -       & -       & -       & -       & -       & -       \\
   amazon & 62.1264 & 63.5019 &   60.8092 & 60.6316 & 60.9868 & -       & -       & -       & -       & -       & -       & -       & -       & -       & -       & -       & -       & -       \\
     MNLI & 61.0796 & 60.4403 &   61.7325 & 52.3158 & 50.3026 & 82.5789 & -       & -       & -       & -       & -       & -       & -       & -       & -       & -       & -       & -       \\
       CB & 61.3035 & 59.9587 &   62.7100 & 54.4474 & 52.3289 & 81.2105 & 82.1429 & -       & -       & -       & -       & -       & -       & -       & -       & -       & -       & -       \\
     COPA & 62.8421 & 60.4403 &   65.4426 & 59.0263 & 58.8816 & 77.9342 & 82.1429 & 93.0000 & -       & -       & -       & -       & -       & -       & -       & -       & -       & -       \\
      QQP & 63.3991 & 59.8899 &   67.3452 & 53.7500 & 52.4605 & 78.7368 & 75.0000 & 93.0000 & 84.0395 & -       & -       & -       & -       & -       & -       & -       & -       & -       \\
      RTE & 62.4633 & 60.5435 &   64.5088 & 57.2500 & 52.6053 & 68.0658 & 83.9286 & 91.0000 & 78.7763 & 87.7256 & -       & -       & -       & -       & -       & -       & -       & -       \\
\bottomrule
\end{tabular}

}
\caption{The task-wise performance (\%) of Baseline under LoRA setting}
\end{table*}

\begin{table*}[htbp]
\centering
\setlength{\tabcolsep}{2pt}
\resizebox{\linewidth}{!}{

\begin{tabular}{lcccccccccccccccccc}
\toprule
Curr.Task &  F1 Avg &    MMLU & Tasks Avg &    yelp &  amazon &    MNLI &      CB &    COPA &     QQP &     RTE &    IMDB &   SST-2 & dbpedia &  agnews &   yahoo & MultiRC &  BoolQA &     WiC \\
\midrule
     yelp & 64.8306 & 65.0843 &   64.5789 & 64.5789 & -       & -       & -       & -       & -       & -       & -       & -       & -       & -       & -       & -       & -       & -       \\
   amazon & 63.1428 & 64.4995 &   61.8421 & 62.4605 & 61.2237 & -       & -       & -       & -       & -       & -       & -       & -       & -       & -       & -       & -       & -       \\
     MNLI & 62.9071 & 63.2611 &   62.5570 & 55.5000 & 54.8684 & 77.3026 & -       & -       & -       & -       & -       & -       & -       & -       & -       & -       & -       & -       \\
       CB & 62.9278 & 61.1283 &   64.8364 & 56.3026 & 55.7763 & 82.2500 & 89.2857 & -       & -       & -       & -       & -       & -       & -       & -       & -       & -       & -       \\
     COPA & 63.3243 & 61.6443 &   65.0984 & 56.8947 & 55.5526 & 82.3026 & 87.5000 & 94.0000 & -       & -       & -       & -       & -       & -       & -       & -       & -       & -       \\
      QQP & 64.3577 & 61.1283 &   67.9474 & 57.6711 & 55.4737 & 75.2368 & 75.0000 & 93.0000 & 83.0263 & -       & -       & -       & -       & -       & -       & -       & -       & -       \\
      RTE & 62.9565 & 59.8555 &   66.3964 & 55.5132 & 53.9605 & 72.8026 & 76.7857 & 90.0000 & 82.1184 & 88.4477 & -       & -       & -       & -       & -       & -       & -       & -       \\
\bottomrule
\end{tabular}

}
\caption{The task-wise performance (\%) of Baseline\textsuperscript{R} under LoRA setting}
\end{table*}

\begin{table*}[htbp]
\centering
\setlength{\tabcolsep}{2pt}
\resizebox{\linewidth}{!}{

\begin{tabular}{lcccccccccccccccccc}
\toprule
Curr.Task &  F1 Avg &    MMLU & Tasks Avg &    yelp &  amazon &    MNLI &      CB &    COPA &     QQP &     RTE &    IMDB &   SST-2 & dbpedia &  agnews &   yahoo & MultiRC &  BoolQA &     WiC \\
\midrule
     yelp & 64.0840 & 64.7747 &   63.4079 & 63.4079 & -       & -       & -       & -       & -       & -       & -       & -       & -       & -       & -       & -       & -       & -       \\
   amazon & 62.9931 & 65.2563 &   60.8816 & 60.6974 & 61.0658 & -       & -       & -       & -       & -       & -       & -       & -       & -       & -       & -       & -       & -       \\
     MNLI & 64.7046 & 63.9491 &   65.4781 & 56.3684 & 56.9868 & 83.0789 & -       & -       & -       & -       & -       & -       & -       & -       & -       & -       & -       & -       \\
       CB & 64.5700 & 63.9147 &   65.2389 & 55.7895 & 57.3421 & 82.4079 & 89.2857 & -       & -       & -       & -       & -       & -       & -       & -       & -       & -       & -       \\
     COPA & 63.7832 & 64.3963 &   63.1817 & 57.6974 & 59.6447 & 71.6184 & 83.9286 & 96.0000 & -       & -       & -       & -       & -       & -       & -       & -       & -       & -       \\
      QQP & 67.3307 & 65.1875 &   69.6197 & 59.8553 & 56.6842 & 77.6184 & 82.1429 & 97.0000 & 83.8684 & -       & -       & -       & -       & -       & -       & -       & -       & -       \\
      RTE & 62.2957 & 63.7083 &   60.9444 & 54.6974 & 47.0658 & 60.6579 & 83.9286 & 98.0000 & 79.6711 & 89.1697 & -       & -       & -       & -       & -       & -       & -       & -       \\
\bottomrule
\end{tabular}

}
\caption{The task-wise performance (\%) of Baseline\textsuperscript{R} with BI under LoRA setting}
\end{table*}

\begin{table*}[htbp]
\centering
\setlength{\tabcolsep}{2pt}
\resizebox{\linewidth}{!}{

\begin{tabular}{lcccccccccccccccccc}
\toprule
Curr.Task &  F1 Avg &    MMLU & Tasks Avg &    yelp &  amazon &    MNLI &      CB &    COPA &     QQP &     RTE &    IMDB &   SST-2 & dbpedia &  agnews &   yahoo & MultiRC &  BoolQA &     WiC \\
\midrule
     yelp & 65.6367 & 66.0819 &   65.1974 & 65.1974 & -       & -       & -       & -       & -       & -       & -       & -       & -       & -       & -       & -       & -       & -       \\
   amazon & 63.7459 & 65.6003 &   61.9934 & 61.8289 & 62.1579 & -       & -       & -       & -       & -       & -       & -       & -       & -       & -       & -       & -       & -       \\
     MNLI & 65.2054 & 64.3963 &   66.0351 & 58.1316 & 56.4605 & 83.5132 & -       & -       & -       & -       & -       & -       & -       & -       & -       & -       & -       & -       \\
       CB & 65.5178 & 63.9147 &   67.2034 & 59.8421 & 58.2763 & 83.3553 & 85.7143 & -       & -       & -       & -       & -       & -       & -       & -       & -       & -       & -       \\
     COPA & 66.0331 & 64.9811 &   67.1197 & 61.0658 & 58.5132 & 81.2368 & 89.2857 & 96.0000 & -       & -       & -       & -       & -       & -       & -       & -       & -       & -       \\
      QQP & 66.7616 & 64.6371 &   69.0306 & 59.5000 & 55.6579 & 76.4737 & 71.4286 & 96.0000 & 84.1184 & -       & -       & -       & -       & -       & -       & -       & -       & -       \\
      RTE & 65.6742 & 64.1211 &   67.3045 & 58.1974 & 53.9342 & 72.6842 & 82.1429 & 95.0000 & 83.1974 & 87.3646 & -       & -       & -       & -       & -       & -       & -       & -       \\
\bottomrule
\end{tabular}

}
\caption{The task-wise performance (\%) of Baseline\textsuperscript{R}+KL under LoRA setting}
\end{table*}

\begin{table*}[htbp]
\centering

\setlength{\tabcolsep}{2pt}
\resizebox{\linewidth}{!}{

\begin{tabular}{lcccccccccccccccccc}
\toprule
Curr.Task &  F1 Avg &    MMLU & Tasks Avg &    yelp &  amazon &    MNLI &      CB &    COPA &     QQP &     RTE &    IMDB &   SST-2 & dbpedia &  agnews &   yahoo & MultiRC &  BoolQA &     WiC \\
\midrule
     yelp & 64.5996 & 65.7379 &   63.5000 & 63.5000 & -       & -       & -       & -       & -       & -       & -       & -       & -       & -       & -       & -       & -       & -       \\
   amazon & 63.6099 & 65.1531 &   62.1382 & 62.5263 & 61.7500 & -       & -       & -       & -       & -       & -       & -       & -       & -       & -       & -       & -       & -       \\
     MNLI & 64.9267 & 64.7059 &   65.1491 & 56.4737 & 56.0526 & 82.9211 & -       & -       & -       & -       & -       & -       & -       & -       & -       & -       & -       & -       \\
       CB & 65.0925 & 64.9467 &   65.2389 & 56.9342 & 56.8816 & 81.6974 & 92.8571 & -       & -       & -       & -       & -       & -       & -       & -       & -       & -       & -       \\
     COPA & 65.4596 & 66.4603 &   64.4886 & 59.4474 & 60.4474 & 72.9605 & 87.5000 & 98.0000 & -       & -       & -       & -       & -       & -       & -       & -       & -       & -       \\
      QQP & 66.9799 & 65.5315 &   68.4939 & 54.2500 & 55.7895 & 79.6579 & 83.9286 & 99.0000 & 83.7632 & -       & -       & -       & -       & -       & -       & -       & -       & -       \\
      RTE & 65.7386 & 65.1875 &   66.2991 & 56.2237 & 54.4474 & 71.5526 & 85.7143 & 97.0000 & 81.6842 & 86.6426 & -       & -       & -       & -       & -       & -       & -       & -       \\
\bottomrule
\end{tabular}

}
\caption{The task-wise performance (\%) of Baseline\textsuperscript{R}+KL with BI under LoRA setting}
\end{table*}

\begin{table*}[htbp]
\centering

\setlength{\tabcolsep}{2pt}
\resizebox{\linewidth}{!}{

\begin{tabular}{lcccccccccccccccccc}
\toprule
Curr.Task &  F1 Avg &    MMLU & Tasks Avg &    yelp &  amazon &    MNLI &      CB &    COPA &     QQP &     RTE &    IMDB &   SST-2 & dbpedia &  agnews &   yahoo & MultiRC &  BoolQA &     WiC \\
\midrule
     yelp & 65.4596 & 66.2539 &   64.6842 & 64.6842 & -       & -       & -       & -       & -       & -       & -       & -       & -       & -       & -       & -       & -       & -       \\
   amazon & 64.1957 & 66.1851 &   62.3224 & 62.8684 & 61.7763 & -       & -       & -       & -       & -       & -       & -       & -       & -       & -       & -       & -       & -       \\
     MNLI & 66.3919 & 65.6347 &   67.1667 & 59.5789 & 58.0132 & 83.9079 & -       & -       & -       & -       & -       & -       & -       & -       & -       & -       & -       & -       \\
       CB & 67.2246 & 66.2883 &   68.1878 & 63.7237 & 60.9211 & 79.7368 & 92.8571 & -       & -       & -       & -       & -       & -       & -       & -       & -       & -       & -       \\
     COPA & 66.6241 & 66.4259 &   66.8235 & 64.6842 & 60.9737 & 74.2763 & 89.2857 & 95.0000 & -       & -       & -       & -       & -       & -       & -       & -       & -       & -       \\
      QQP & 68.4241 & 66.3915 &   70.5852 & 62.7632 & 58.2632 & 77.2500 & 83.9286 & 97.0000 & 83.6184 & -       & -       & -       & -       & -       & -       & -       & -       & -       \\
      RTE & 67.2146 & 65.4283 &   69.1013 & 60.9474 & 55.7500 & 75.7895 & 85.7143 & 92.0000 & 82.7895 & 88.4477 & -       & -       & -       & -       & -       & -       & -       & -       \\
\bottomrule
\end{tabular}

}
\caption{The task-wise performance (\%) of Baseline\textsuperscript{R}+L1 under LoRA setting}
\end{table*}

\begin{table*}[htbp]
\centering

\setlength{\tabcolsep}{2pt}
\resizebox{\linewidth}{!}{

\begin{tabular}{lcccccccccccccccccc}
\toprule
Curr.Task &  F1 Avg &    MMLU & Tasks Avg &    yelp &  amazon &    MNLI &      CB &    COPA &     QQP &     RTE &    IMDB &   SST-2 & dbpedia &  agnews &   yahoo & MultiRC &  BoolQA &     WiC \\
\midrule
     yelp & 64.7243 & 65.8411 &   63.6447 & 63.6447 & -       & -       & -       & -       & -       & -       & -       & -       & -       & -       & -       & -       & -       & -       \\
   amazon & 64.0712 & 66.1507 &   62.1184 & 62.5000 & 61.7368 & -       & -       & -       & -       & -       & -       & -       & -       & -       & -       & -       & -       & -       \\
     MNLI & 66.7418 & 66.0131 &   67.4868 & 58.8158 & 60.5789 & 83.0658 & -       & -       & -       & -       & -       & -       & -       & -       & -       & -       & -       & -       \\
       CB & 66.1626 & 65.5659 &   66.7702 & 59.2763 & 60.6579 & 80.1974 & 91.0714 & -       & -       & -       & -       & -       & -       & -       & -       & -       & -       & -       \\
     COPA & 65.0898 & 66.3571 &   63.8700 & 60.0395 & 59.8816 & 71.0395 & 91.0714 & 98.0000 & -       & -       & -       & -       & -       & -       & -       & -       & -       & -       \\
      QQP & 66.9136 & 65.9787 &   67.8754 & 56.5000 & 55.9474 & 76.0658 & 71.4286 & 94.0000 & 82.6184 & -       & -       & -       & -       & -       & -       & -       & -       & -       \\
      RTE & 64.9409 & 65.6003 &   64.2947 & 59.5921 & 57.1974 & 61.9342 & 80.3571 & 95.0000 & 77.0526 & 88.4477 & -       & -       & -       & -       & -       & -       & -       & -       \\
\bottomrule
\end{tabular}

}
\caption{The task-wise performance (\%) of Baseline\textsuperscript{R}+L1 with BI under LoRA setting}
\end{table*}

\begin{table*}[htbp]
\centering

\setlength{\tabcolsep}{2pt}
\resizebox{\linewidth}{!}{

%

}
\caption{The task-wise performance (\%) of Baseline\textsuperscript{R}+L2 under LoRA setting}
\end{table*}

\begin{table*}[htbp]
\centering

\setlength{\tabcolsep}{2pt}
\resizebox{\linewidth}{!}{

%

}
\caption{The task-wise performance (\%) of Baseline\textsuperscript{R}+L2 with BI under LoRA setting}
\end{table*}

\begin{table*}[htbp]
\centering

\setlength{\tabcolsep}{2pt}
\resizebox{\linewidth}{!}{

%

}
\caption{The task-wise performance (\%) of Baseline\textsuperscript{R}+TM under LoRA setting}
\end{table*}

\begin{table*}[htbp]
\centering

\setlength{\tabcolsep}{2pt}
\resizebox{\linewidth}{!}{

%

}
\caption{The task-wise performance (\%) of Baseline\textsuperscript{R}+TM with BI under LoRA setting}
\end{table*}

\begin{table*}[htbp]
\centering

\setlength{\tabcolsep}{2pt}
\resizebox{\linewidth}{!}{

%

}
\caption{The task-wise performance (\%) of Baseline\textsuperscript{R}+TM (w=100) under LoRA setting}
\end{table*}

\begin{table*}[htbp]
\centering

\setlength{\tabcolsep}{2pt}
\resizebox{\linewidth}{!}{

%

}
\caption{The task-wise performance (\%) of Baseline\textsuperscript{R}+TM (w=100) with BI under LoRA setting}
\end{table*}

\begin{table*}[htbp]
\centering

\setlength{\tabcolsep}{2pt}
\resizebox{\linewidth}{!}{

%

}
\caption{The task-wise performance (\%) of Baseline\textsuperscript{R}+TM (w=dynamic) under LoRA setting}
\end{table*}

\begin{table*}[htbp]
\centering

\setlength{\tabcolsep}{2pt}
\resizebox{\linewidth}{!}{

%

}
\caption{The task-wise performance (\%) of Baseline\textsuperscript{R}+TM (w=dynamic) with BI under LoRA setting}
\label{tab:detail_lora_last}
\end{table*}

\begin{table*}[htbp]
\centering

\setlength{\tabcolsep}{2pt}
\resizebox{\linewidth}{!}{

%

}
\caption{The task-wise performance (\%) of O-LoRA}
\label{tab:detail_olora}
\end{table*}

\end{document}